\pdfoutput=1

\documentclass[11pt]{article}

\usepackage{acl}

\usepackage{times}
\usepackage{latexsym}

\usepackage[T1]{fontenc}

\usepackage[utf8]{inputenc}

\usepackage{microtype}
\newcommand*\samethanks[1][\value{footnote}]{\footnotemark[#1]}
\usepackage{hyperref}
\usepackage{xcolor}
\usepackage{tabularx}
\usepackage{multirow}
\usepackage{graphicx}
\usepackage[frozencache ,cachedir=.]{minted}
\usepackage{booktabs}

\usepackage{amsmath}
\usepackage{multirow}
\usepackage{pgfplots}
\usepackage{pgfplotstable}
\usepackage{subcaption}

\title{Towards Efficient NLP: A Standard Evaluation and A Strong Baseline}

\author{
Xiangyang Liu\textsuperscript{\rm 1,2}\thanks{{} {} Equal contribution. }\ ,
Tianxiang Sun\textsuperscript{\rm 1,2}\samethanks\ ,
Junliang He\textsuperscript{\rm 1,2}, 
Jiawen Wu\textsuperscript{\rm 1,2}, 
Lingling Wu\textsuperscript{\rm 1,2},\\
\bf{
Xinyu Zhang\textsuperscript{\rm 3}, 
Hao Jiang\textsuperscript{\rm 3}, 
Zhao Cao\textsuperscript{\rm 3}, 
Xuanjing Huang\textsuperscript{\rm 1,2}, 
Xipeng Qiu\textsuperscript{\rm 1,2 \thanks{{} {} Corresponding author.}}
}\\
\textsuperscript{\rm 1}School of Computer Science, Fudan University\\
\textsuperscript{\rm 2}Shanghai Key Laboratory of Intelligent Information Processing, Fudan University\\
\textsuperscript{\rm 3}Huawei Poisson Lab\\
\tt \{xiangyangliu20,txsun19,xjhuang,xpqiu\}@fudan.edu.cn\\
\tt \{zhangxinyu35,jianghao66,caozhao1\}@huawei.com \\
}

\pgfplotsset{compat=1.17}
\begin{document}
\maketitle
\begin{abstract}
Supersized pre-trained language models have pushed the accuracy of various natural language processing (NLP) tasks to a new state-of-the-art (SOTA). Rather than pursuing the reachless SOTA accuracy, more and more researchers start paying attention on model efficiency and usability. Different from accuracy, the metric for efficiency varies across different studies, making them hard to be fairly compared. To that end, this work presents \textbf{ELUE} (\textbf{E}fficient \textbf{L}anguage \textbf{U}nderstanding \textbf{E}valuation), a standard evaluation, and a public leaderboard for efficient NLP models. ELUE is dedicated to depict the Pareto Frontier for various language understanding tasks, such that it can tell whether and how much a method achieves Pareto improvement. Along with the benchmark, we also release a strong baseline, \textbf{ElasticBERT}, which allows BERT to exit at any layer in both static and dynamic ways. We demonstrate the ElasticBERT, despite its simplicity, outperforms or performs on par with SOTA compressed and early exiting models. With ElasticBERT, the proposed ELUE has a strong Pareto Frontier and makes a better evaluation for efficient NLP models.
\end{abstract}

\section{Introduction}
Driven by the large-scale pre-training, today's NLP models have become much more powerful~\citep{Devlin2019BERT,Yang2019XLNet,Lan2020ALBERT,Raffel2020T5,Sun2020Colake,Brown2020GPT3,Qiu2020survey}. As a consequence of this drastic increase in performance, these pre-trained language models (PLMs) are notorious for becoming more and more computationally expensive due to the increasing number of parameters. Therefore, rather than pre-training a larger model to achieve a new state-of-the-art (SOTA) accuracy, 
most studies are pursuing improvement on other dimensions such as the number of parameters or FLOPs~\citep{Gordon2020Compressing,Sanh2019DistilBERT,Jiao2020TinyBERT,Lan2020ALBERT,Shen2020QBERT}. For these works, the goal has shifted from simple SOTA to "Pareto SOTA". A Pareto SOTA model means that there is no other model is currently better than it on all the dimensions of interest. For example, a model may claim to be Pareto SOTA as long as it achieves the best accuracy under the same number of parameters or FLOPs. For these efficient models with fewer parameters or FLOPs, it is unfair to get them evaluated on the accuracy-centric benchmarks such as GLUE~\cite{Wang2019GLUE}, and ranked among many large-scale models.

\begin{figure}[t!]
    \centering
    \includegraphics[width=\linewidth]{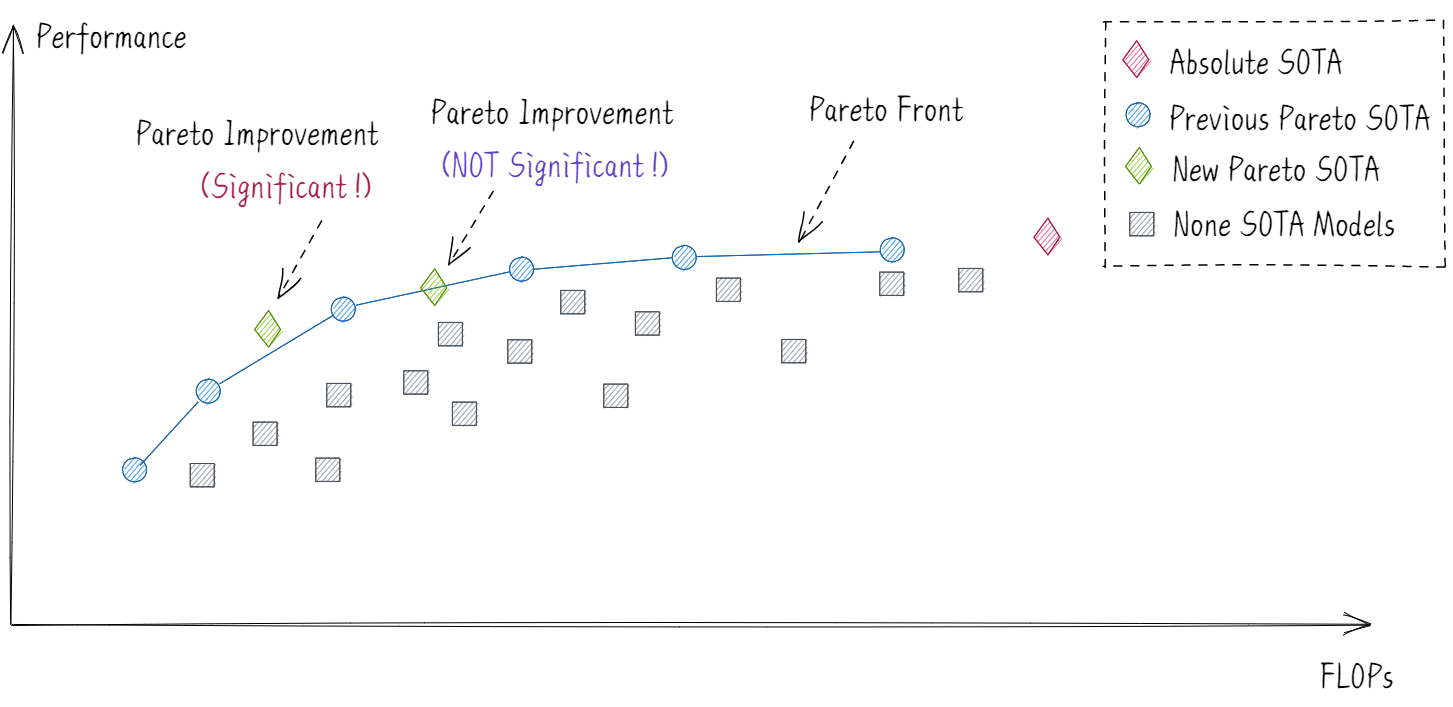}
    \caption{An illustration to show our motivation, that is, building the Pareto frontier can help recognizing whether and how much a method achieves Pareto improvement.\vspace{-0.5cm}}
    \label{fig:pareto}
\end{figure}

The shifted goal has outpaced the existing benchmarks, which cannot provide a comprehensive and intuitive comparison for efficient methods. In the absence of a proper benchmark, measures of efficiency in different studies cannot be standardized, and different methods cannot be fairly compared. As a result, it is difficult to say \textit{whether} and \textit{how much} a method achieves Pareto improvement. To that end, we aim to build the Pareto frontier for various tasks with standard evaluation for both performance and efficiency. Our motivation can be briefly illustrated by Figure~\ref{fig:pareto}. 

\paragraph{Need for a standard evaluation}
As the goal has shifted, a new benchmark is urgently needed to comprehensively compare the NLP models in multiple dimensions. Currently, this multi-dimensional comparison is done in the individual papers, resulting in the following issues: \textbf{(a) Incomprehensive comparison.} The comparison is usually point-to-point, e.g. comparing model performance under the same FLOPs. The comparison in a broader range is usually missed, especially for works in conditional computation where the model performance varies with FLOPs. \textbf{(b) Unaccessible results.} Even if the comprehensive line-to-line comparison is conducted, the results are usually presented in form of figure, in which the data points are not accessible for the following work. As a result, the following work has to reproduce or estimate the results (e.g. \citet{Xin2021BERxiT} estimate values from the figures of \citet{Zhou2020PABEE}). \textbf{(c) Non-standard measurements.} Different works may adopt different metrics such as physical elapsed time, FLOPs, and executed model layers, making them hard to directly compare. Even if the adopted metrics are the same, there is no guarantee that they will be calculated in the same way (e.g. the hardware infrastructure, or the software to calculate FLOPs can be very different\footnote{We find that the FLOPs of Transformers calculated by different libraries (\texttt{thop}, \texttt{ptflops}, and \texttt{torchstat}) can be different. And besides, all of them missed FLOPs in some operations such as self-attention and layer normalization.}). \textbf{(d) Inconvenience.} Recent studies usually choose GLUE~\cite{Wang2019GLUE} as the main benchmark, which, however, is not suitable for dynamic methods due to its submission limitation that is designed to avoid overfitting on test sets.

\paragraph{Need for a strong baseline}
Currently, there are roughly two branches of efficient methods in NLP: static methods (e.g. distillation, pruning, quantization, etc.) and dynamic methods (e.g. early exiting). \textbf{(a) Static models} are obtained given an expected number of parameters or inference latency. These methods often use the first few layers (to keep the same number of parameters or FLOPs) of some pre-trained model followed by a classification head as their baseline, which, however, is too weak to serve as a baseline. \textbf{(b) Dynamic models} usually add multiple internal classifiers to the pre-trained LMs, and therefore allow flexible inference conditioned on the input. Nevertheless, the injected internal classifiers introduce a gap between pre-training and fine-tuning. Training the internal classifiers on downstream tasks often degenerates the performance of the entire model~\cite{Xin2021BERxiT}. Thus, static models need a strong baseline, and dynamic models need a strong backbone.


\paragraph{Contributions} In this work, we address the above needs by contributing the following:

\begin{itemize}
    \item \textbf{ELUE}(\textbf{E}fficient \textbf{L}anguage \textbf{U}nderstanding \textbf{E}valuation) -- a standard benchmark for efficient NLP models. (1) ELUE supports online evaluation for model performance, FLOPs, and number of parameters. (2) ELUE is also an open-source platform that can facilitate future research. We reproduce and evaluate multiple compressed and early exiting methods on ELUE. All of the results are publicly accessible on ELUE. (3) ELUE provides an online leaderboard that uses a specific metric to measure how much a model oversteps the current Pareto frontier. ELUE leaderboard also maintains several separate tracks for models with different sizes. (4) ELUE covers six NLP datasets spanning sentiment analysis, natural language inference, similarity and paraphrase tasks. The ELUE benchmark is publicly available at~\url{http://eluebenchmark.fastnlp.top/}.
    \item \textbf{ElasticBERT} -- a strong baseline (backbone) for static (dynamic) models. ElasticBERT is a multi-exit Transformer~\cite{Vaswani2017Attention} pre-trained on $\sim$160GB corpus. The pre-training objectives, MLM and SOP~\cite{Lan2020ALBERT}, are applied to multiple Transformer layers instead of only the last layer. Gradient equilibrium~\cite{Li2019Improve} is adopted to alleviate the conflict of the losses at different layers. For static models, ElasticBERT is a strong baseline that can reach or even outperform distilled models. For dynamic models, ElasticBERT is a robust backbone that closes the gap between pre-training and fine-tuning. We release the pre-trained model weights of ElasticBERT\footnote{\href{https://huggingface.co/fnlp}{https://huggingface.co/fnlp}} as well as code\footnote{\href{https://github.com/fastnlp/ElasticBERT}{https://github.com/fastnlp/ElasticBERT}}.
\end{itemize}

\section{Related Work}
\paragraph{NLP Benchmarks}
Evaluating the quality of language representations on multiple downstream tasks has become a common practice in the community. These evaluations have measured and pushed the progress of NLP in recent years. SentEval~\cite{Conneau2018SentEval} introduces a standard evaluation toolkit for multiple NLP tasks.
Further, GLUE~\cite{Wang2019GLUE} and SuperGLUE~\cite{Wang2019Superglue} provide a set of more difficult datasets for model-agnostic evaluation. Another line of work is multi-dimensional evaluations. EfficientQA~\cite{Min2020EfficientQA} is an open-domain question answering challenge that evaluates both accuracy and system size. The system size is measured as the number of bytes required to store a Docker image that contains the submitted system. Dynabench~\cite{Kiela2021Dynabench}, an open-source benchmark for dynamic dataset creation and model evaluation, also supports multi-dimensional evaluation. In particular, Dynabench measures model performance, throughput, memory use, fairness, and robustness. Both EfficientQA and Dynabench require the user to upload the model along with the required environment to the server, which is costly for users to upload and also for the server to evaluate. In contrast, ELUE adopts a cheaper way to evaluate performance and efficiency of the model. Recently, Long-Range Arena (LRA)~\cite{Tay2021LRA} is proposed to evaluate models under the long-context scenario. Different from ELUE, LRA mainly focuses on Xformers~\cite{lin2021transformers}. Besides, some tasks included in LRA are not NLP tasks, or even not real-world tasks, while ELUE consists of common language understanding tasks. In addition, ELUE is also inspired by other well-known benchmarks, such as SQuAD~\cite{Rajpurkar2016SQuAD}, MultiNLI~\cite{Williams2018MNLI}, DecaNLP~\cite{McCann2018DecaNLP}, CLUE~\cite{Xu2020CLUE}, HotpotQA~\cite{Yang2018HotpotQA}, GEM~\cite{gem2021Gehrmann}, etc.

\paragraph{Efficient NLP Models} Current efficient NLP models can be roughly categorized as two streams: model compression (static methods) and conditional computation (dynamic methods). Model compression is to reduce the number or precision of model parameters to achieve faster training and inference. Currently, there are several ways to achieve model compression: (1) \textit{Knowledge Distillation}, which is to learn a compact student model that learns from the output distribution of a large-scale teacher model~\cite{Sanh2019DistilBERT,Jiao2020TinyBERT} (2) \textit{Model Pruning}, which is to remove parts of parameters that are less important~\cite{Gordon2020Compressing}, (3) \textit{Weight Sharing} across different parts of the model~\cite{Lan2020ALBERT} is also a common technique to significantly reduce parameters, (4) \textit{Quantization}, which is to use low bit precision to store parameter and accelerate inference with low bit hardware operations~\cite{Shen2020QBERT}, and (5) \textit{Module Replacing}, which is to replace the modules of a big model with more compact substitutes~\cite{Xu2020BERTTheseus}. In contrast, conditional computation is to selectively execute only parts of the model conditioned on a given input~\cite{Bengio2013Estimating,Davis2013Lowrank}. As a representative, an end-to-end halting approach, Adaptive Computation Time (ACT)~\cite{Graves2016Adaptive}, is developed to perform input-adaptive computation for recurrent networks. The idea of ACT is later adopted in Universal Transformer~\cite{Dehghani2019Universal}. Recently, as the rising of deep models for natural language processing, early exiting is widely used to speedup inference of transformer models~\cite{Liu2020FastBERT,Xin2020DeeBERT,Schwartz2020Right,Zhou2020BERT,Elbayad2020Depth,Liao2021Global,Xin2021BERxiT,Sun2021Early,Zhu2021Leebert,Li2021cascadebert}.

\section{ELUE: A Standard Benchmark for Efficient NLP Models}
ELUE aims to offer a standard evaluation for various efficient NLP models, such that they can be fairly and comprehensively compared. In Section~\ref{sec:design}, we list the design considerations to achieve this motivation. In Section~\ref{sec:task}, we describe the tasks and datasets included in ELUE. In Section~\ref{sec:eval}, we illustrate how to make a submission on ELUE, and how the submission is evaluated. In Section~\ref{sec:leaderboard}, we discuss the design of our leaderboard.

\subsection{Design Considerations}
\label{sec:design}
Now we enumerate main considerations in the design of ELUE to ensure that it meets the needs mentioned early.
\paragraph{Multi-dimensional Evaluation}
The evaluation of ELUE should be multi-dimensional for comprehensive comparison. Instead of point-to-point comparison, methods can be compared in a line-to-line style in ELUE, where the "line" is a performance-efficiency trade-off curve.

\paragraph{Public Accessible}
All data points in ELUE should be publicly accessible such that the following work does not need to reproduce or estimate results from previous work. To facilitate future research, some representative methods should be reproduced and evaluated in ELUE.

\paragraph{Standard Evaluation}
The measurement of model efficiency should be standardized in ELUE such that this line of methods can be fairly compared. Current studies usually use number of parameters~\cite{Lan2020ALBERT,Jiao2020TinyBERT}, FLOPs~\cite{Jiao2020TinyBERT,Liu2020FastBERT,Li2020Accelerating}, actual inference time~\cite{Sanh2019DistilBERT,Schwartz2020Right}, or number of executed layers~\cite{Zhou2020PABEE,Sun2021Early} to measure model efficiency. Among these metrics, measuring actual inference time is costly for both users and the server, and highly depends on the computation infrastructure and software implementation, while number of executed layers ignores the shape of input and hidden layers, therefore is inaccurate. Thus, ELUE adopts number of parameters and FLOPs as the metrics for model efficiency.

\paragraph{Easy-to-Use}
ELUE should be friendly to users, which means that the submission should be as simple as possible. Roughly speaking, there are currently two ways of submissions: (1) submitting the trained model such as SQuAD~\cite{Rajpurkar2016SQuAD}, Dynabench~\cite{Kiela2021Dynabench}, and (2) submitting the predicted test files such as GLUE~\cite{Wang2019GLUE}, SuperGLUE~\cite{Wang2019Superglue}, and CLUE~\cite{Xu2020CLUE}. The submission of ELUE lies in the latter way. Nevertheless, to evaluate number of parameters and FLOPs, the submitted test files should conform to a specific format, and besides, a Python file to define the used model is also required. For more details about submission and evaluation, see Appendix~\ref{sec:eval}.

\subsection{Task and Dataset Selection}
\label{sec:task}
Following GLUE~\cite{Wang2019GLUE}, SuperGLUE~\cite{Wang2019Superglue}, and CLUE~\cite{Xu2020CLUE}, we collect tasks that can be formatted as single sentence classification or sentence pair classification. Since ELUE mainly focuses on efficient models, the difficulty of dataset is not a primary consideration. Instead, we collect tasks and datasets that are commonly used and publicly available in the community. The statistics of the collected datasets are listed in Table~\ref{tab:dataset}.

\begin{table}[h]
\centering
\resizebox{\linewidth}{!}{
\begin{tabular}{llrrr}
\toprule
\textbf{Tasks}                                                                          & \textbf{Datasets} & \textbf{|Train|} & \textbf{|Dev|} & \textbf{|Test|} \\ \midrule
\multirow{2}{*}{\begin{tabular}[c]{@{}l@{}}Sentiment\\ Analysis\end{tabular}}           & SST-2             & 8,544            & 1,101          & 2,208           \\
                                                                                        & IMDb              & 20,000           & 5,000          & 25,000          \\ \midrule
\multirow{2}{*}{\begin{tabular}[c]{@{}l@{}}Natural Language\\ Inference\end{tabular}} & SNLI              & 549,367          & 9,842         & 9,824          \\
                                                                                        & SciTail           & 23,596           & 1,304          & 2,126           \\ \midrule
\multirow{2}{*}{\begin{tabular}[c]{@{}l@{}}Similarity and\\ Paraphrase\end{tabular}}  & MRPC              & 3,668            & 408            & 1,725           \\
                                                                                        & STS-B             & 5,749            & 1,500          & 1,379           \\ \bottomrule
\end{tabular}
}
\caption{Statistics of datasets in ELUE.\vspace{-0.5cm}}
\label{tab:dataset}
\end{table}

\paragraph{Sentiment Analysis}
Sentiment analysis, which is to classify the polarity of a given text, is a fundamental task in NLP. We select two well-known movie review datasets, Stanford Sentiment Treebank (SST)~\cite{Socher2013SST} and IMDb~\cite{Maas2011IMDb}. For SST, we use the two-way class split, i.e. SST-2. Different from GLUE, SST-2 samples in ELUE are complete sentences instead of phrases. For IMDb, we randomly select 2.5k positive samples and 2.5k negative samples from training set to construct a development set.

\paragraph{Natural Language Inference}
Natural language inference (NLI) is a task to predict whether the premise entails the hypothesis, contradicts the hypothesis, or neither. NLI is often formulated as a sentence pair classification task~\cite{Devlin2019BERT,Sun2021Paradigm}. We select two NLI datasets, SNLI~\cite{Bowman2015SNLI} and SciTail~\cite{Khot2018SciTail}. SNLI is a crowd-sourced collection of sentence pairs with balanced labels: \textit{entailment}, \textit{contradiction}, and \textit{neutral}. We use the spell-checked version of the test and development sets\footnote{\href{https://nlp.stanford.edu/projects/snli/}{https://nlp.stanford.edu/projects/snli/}}. The hard samples, which do not have golden labels due to the disagreement of annotators, are removed from the dataset and left for model diagnostic. SciTail is a two-way (\textit{entail} or \textit{neutral}) entailment classification dataset, which is derived from multiple-choice science exams and web sentences.

\paragraph{Similarity and Paraphrase}
For similarity and paraphrase tasks, we also select two datasets, Microsoft Research Paraphrase Corpus (MRPC)~\cite{Dolan2005MRPC}, and Semantic Textual Similarity Benchmark (STS-B)~\cite{Cer2017SemEval2017}, both of which are also included in GLUE. MRPC is a collection of automatically extracted sentence pairs, each manually-labeled with a judgment to indicate whether the pair constitutes a paraphrase. STS-B is a corpus of sentence pairs, each of which is labeled with a score from 0 to 5 to represent the degree to which two sentences are semantically equivalent.

\begin{figure}[t]
    \centering
    \includegraphics[width=\linewidth]{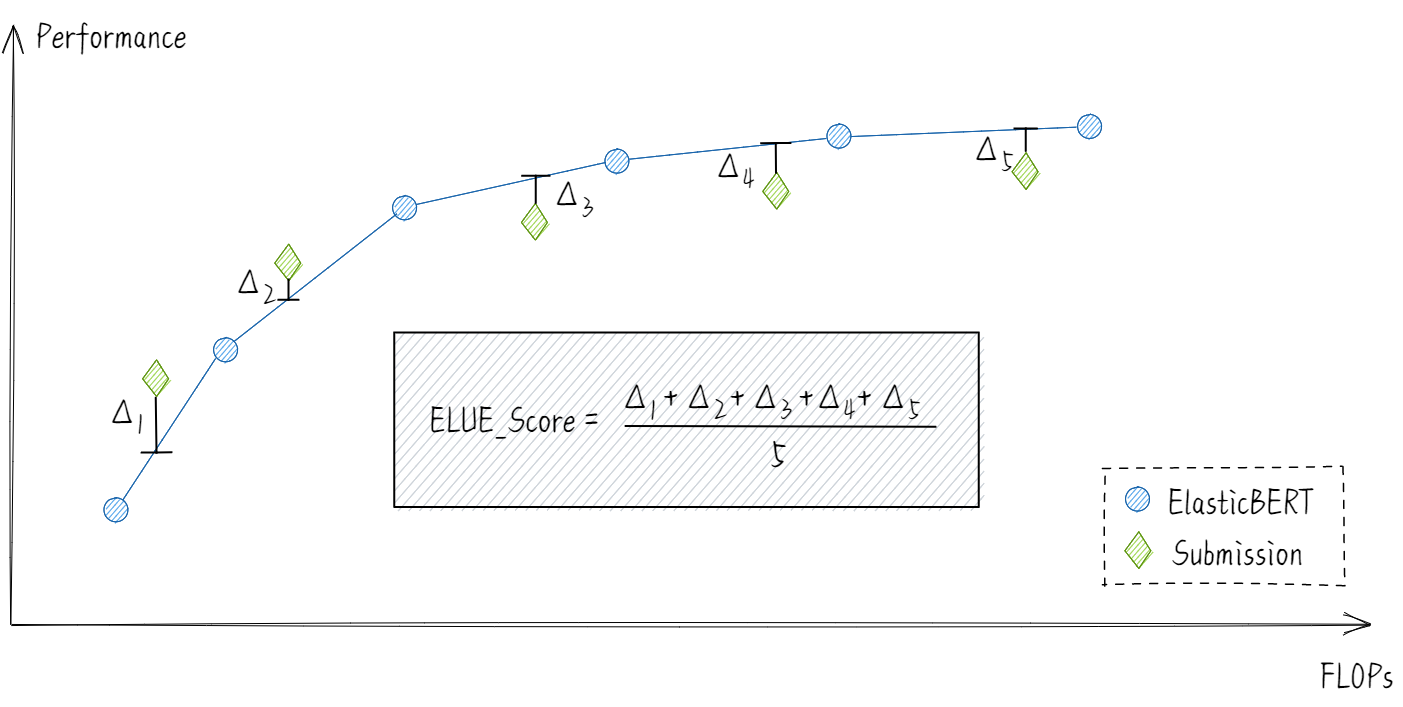}
    \caption{An illustration to show how ELUE score is computed.\vspace{-0.5cm}}
    \label{fig:elue_score}
\end{figure}

\subsection{Leaderboard}
\label{sec:leaderboard}
Following prior work~\cite{Yang2018HotpotQA,Wang2019GLUE,Xu2020CLUE}, we also integrate a leaderboard in ELUE. For dynamic models that have multiple performance-FLOPs coordinates on each dataset, we need to sum up these coordinates as a score. A critical problem is to measure how good a coordinate is. In other words, to measure a coordinate $(p, f)$, where $p$ is performance and $f$ is FLOPs, we need a baseline performance under the same FLOPs. We choose ElasticBERT as the baseline curve. We evaluate different layers of ElasticBERT, and obtained 12 coordinates $(p^{EB}_i, f^{EB}_i)_{i=1}^{12}$, which are then used to interpolate to get a performance-FLOPs function $p^{EB}(f)$. With the baseline curve at hand, we can score a submission curve as
\begin{equation}
    \text{ELUEScore} = \frac{1}{n}\sum_{i=1}^n [p_i - p^{EB}(f_i)].
\label{eq:elue_score}
\end{equation}
Note that the coordinates of ElasticBERT are separately interpolated on different datasets. The final ELUE score is an unweighted average of the scores on all the 6 datasets. Figure~\ref{fig:elue_score} gives an illustration of how ELUE score is computed. The ELUE score reflects the extent to which the submission oversteps the ElasticBERT.

In addition, following EfficientQA~\cite{Min2020EfficientQA}, ELUE leaderboard also maintains four additional separate tracks, corresponding to models below 40M, 55M, 70M, 110M parameters. Models in these tracks are ranked by the average performance on all the datasets.

\begin{figure*}[t!]
    \centering
    \includegraphics[width=\linewidth]{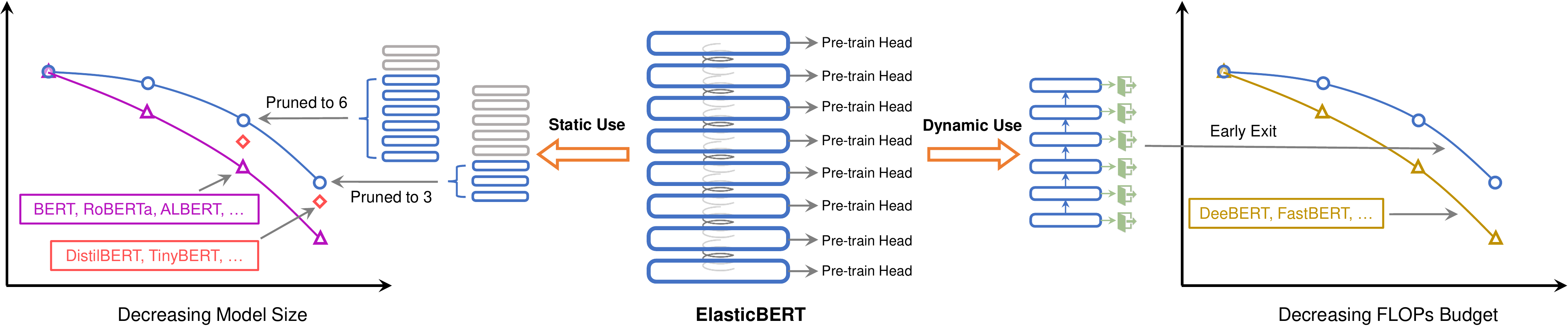}
    \caption{ElasticBERT is pre-trained with multiple pre-training heads attached at the intermediate layers. For static usage (left), it can be pruned on demand while outperforming previous pre-trained models with the same size. For dynamic usage (right), it can serve as the backbone for early exiting methods, achieving better performance-efficiency trade-off than early exiting models with other backbones.}
    \label{fig:elasticity of elasticbert}
\end{figure*}

\section{ElasticBERT: A Strong Baseline for Efficient Inference}
Despite the encouraging results achieved by existing efficient models, we argue that a strong baseline (backbone) is needed for both static methods and dynamic methods. Static methods often choose the first few layers of some pre-trained models as their baseline (e.g. \citet{Sun2019PKD,Jiao2020TinyBERT}), which can be weak. Dynamic methods that enable early exiting by training multiple internal classifiers usually introduce a gap between pre-training and fine-tuning, and therefore hurt the performance of the entire model~\cite{Xin2021BERxiT}. Thus, as illustrated in Figure ~\ref{fig:elasticity of elasticbert}, we present the ElasticBERT that bridges the gap between static and dynamic methods, and therefore can serve as a strong baseline for static methods and also a strong backbone for dynamic methods.

ElasticBERT is a multi-exit pre-trained language model with the following training objective:
\begin{equation}
    \mathcal{L} = \sum_{l=1}^{L}(\mathcal{L}^{\text{MLM}}_l + \mathcal{L}^{\text{SOP}}_l),
    \label{eq:loss}
\end{equation}
where $L$ is the total number of layers, $\mathcal{L}^{\text{MLM}}$ is the n-gram masked language modeling loss, $\mathcal{L}^{\text{SOP}}$ is the sentence order prediction loss~\cite{Lan2020ALBERT}. The two losses are applied to each layer of the model, such that the number of layers can be flexibly scaled on downstream tasks, and therefore it is named "ElasticBERT".

\paragraph{Bridge the Gap Between Static and Dynamic Methods}
As a baseline for static methods, the depth of ElasticBERT can be flexibly reduced on demand. Compared with the first $l$ layer of BERT~\cite{Devlin2019BERT}, the $l$-layered ElasticBERT is a complete model~\cite{Turc2019BERTComplete,Li2021cascadebert} and can achieve better performance. It is worth noticing that ElasticBERT can be regarded as a special instance of LayerDrop~\cite{Fan2020LayerDrop} where the dropped layers are constrained to the top consecutive layers. As a backbone for dynamic methods, training classifiers injected in intermediate layers would be consistent with pre-training. Therefore, ElasticBERT can not only be used as a static complete model, but also be used as a backbone model of dynamic early exiting.

\paragraph{Gradient Equilibrium}
Pre-training with the simply summed loss in Eq. (\ref{eq:loss}) could lead to a \textit{gradient imbalance} issue~\cite{Li2019Improve}. In particular, due to the overlap of subnetworks, the variance of the gradient may grow overly large, leading to unstable training. To address this issue, we follow \citet{Li2019Improve} and adopt the gradient equilibrium (GE) strategy\footnote{The reader is referred to the original paper for more details. In brief, the gradients of $\mathcal{L}_j$ w.r.t. the parameters of the $i$-th layer ($i<j$) would be properly rescaled.} in the pre-training of ElasticBERT.

\paragraph{Grouped Training}
In our preliminary experiments, we found that summing up losses at all layers could slow down pre-training and increase memory footprints. To alleviate this, we divide $L$ exits into $G$ groups. During training, we optimize the losses of the exits within each group by cycling alternately between different batches:
\begin{equation}
    \mathcal{L} = \sum_{l\in \mathcal{G}_i} (\mathcal{L}^{\text{MLM}}_l + \mathcal{L}^{\text{SOP}}_l).
\label{eq:group}
\end{equation}
In Section~\ref{sec:ablation} we explore the performance of different grouping methods. As a result, we group the 12 exits of ElasticBERT\textsubscript{BASE} into $\mathcal{G}_1$=\{1, 3, 5, 7, 9, 11, 12\} and $\mathcal{G}_2$=\{2, 4, 6, 8, 10, 12\}, and group the 24 exits of ElasticBERT\textsubscript{LARGE} into $\mathcal{G}_1$=\{1, 4, 7, ..., 22, 24\}, $\mathcal{G}_2$=\{2, 5, 8, ..., 23, 24\}, and $\mathcal{G}_3$=\{3, 6, 9, ..., 21, 24\}. Our experiments demonstrate that grouped training can significantly speedup the process of pre-training without a loss in performance.



\section{Experiments}
\label{sec:exp}

\subsection{Experimental Setup}
\paragraph{Pre-training Setup} 
Following BERT~\cite{Devlin2019BERT}, we train ElasticBERT in two different configurations: ElasticBERT\textsubscript{BASE} and ElasticBERT\textsubscript{LARGE}, which have the same model sizes with BERT\textsubscript{BASE} and BERT\textsubscript{LARGE}, respectively. The detailed description can be found in Appendix  \ref{sec:trainingsetup}.

\paragraph{Downstream Evaluation}
We evaluate ElasticBERT on the ELUE benchmark, as a static model and as a dynamic model. As a static model, we evaluate different layers of ElasticBERT, denoted as ElasticBERT-$n$L. As a dynamic model, we inject and train internal classifiers in ElasticBERT\textsubscript{BASE} and adopt two strategies, entropy~\cite{Xin2020DeeBERT} and patience~\cite{Zhou2020PABEE}, to enable early exiting, denoted as ElasticBERT\textsubscript{entropy} and ElasticBERT\textsubscript{patience}. To compare with previous work, we also evaluate ElasticBERT on the GLUE benchmark~\cite{Wang2019GLUE}. The comparison results is shown in Appendix  \ref{sec:elasticBERTonGLUE}. For static usage, we fine-tune ElasticBERT and our baseline models for 10 epochs with early stopping using AdamW optimizer~\cite{Ilya2019AdamW} with learning rates of \{1e-5, 2e-5, 3e-5\} and batch size of 32, and warm up the learning rate for the first 6 percent of total steps. In dynamic usage, for the models using two-stage training methods, we train for 3 epochs for each stage, and we train for 5 epochs for other models. Other optimization configurations are the same as those in static scenario.   


\begin{table*}[t!]
\centering
\resizebox{.95\linewidth}{!}{
\begin{tabular}{lccccccccc}
\toprule
\textbf{Model} & \textbf{\#Params} & \textbf{\#FLOPs} & \textbf{SST-2} & \textbf{IMDb} & \textbf{MRPC} & \textbf{STS-B} & \textbf{SNLI} & \textbf{SciTail} & \textbf{Average} \\
\midrule
\multicolumn{10}{c}{\textit{BASE Models}} \\
\midrule
BERT\textsubscript{BASE} & 109M & 13399M & 85.1 & 93.0 & 83.1 & 84.2 & 90.4 & 93.2 & 88.2 \\  
ALBERT\textsubscript{BASE} & 12M & 13927M & 86.6 & 92.9 & 87.8 & 88.3 & 90.1 & 93.4 & 89.9 \\
RoBERTa\textsubscript{BASE} & 125M & 13103M & 88.3 & \textbf{94.9} & 88.0 & \textbf{89.6} & 91.3 & 92.8 & \textbf{90.8} \\
LayerDrop\textsubscript{BASE} & 125M & 13103M & 88.5 & 94.2 &  \textbf{88.2} & 87.1 & 90.7 & 92.8 & 90.3 \\
\textbf{ElasticBERT}\textsubscript{BASE} & 109M & 13399M & \textbf{88.6} & 93.9 & 87.9 & 87.6 & \textbf{91.3} & \textbf{93.8} & 90.5 \\
\midrule
BERT\textsubscript{BASE}-6L & 67M & 6700M & 83.3 & 91.0 & 82.6 & 82.5 & 88.9 & 90.7 & 86.5 \\  
ALBERT\textsubscript{BASE}-6L & 12M & 6972M & 84.7 & 92.0 & 85.3 & 83.5 & 89.3 & 92.3 & 87.9 \\
RoBERTa\textsubscript{BASE}-6L & 82M & 6552M & 86.8 & 92.6 & 86.7 & 84.5 & \textbf{90.2} & 91.3 & 88.7 \\
LayerDrop\textsubscript{BASE}-6L & 82M & 6552M & 86.3 & \textbf{92.9} & 86.3 & 86.1 & 89.5 & 90.3 & 88.6 \\
HeadPrune-BERT\textsubscript{BASE} & 86M & 9249M & 84.8 & 84.7 & 77.8 & 74.8 & 87.8 & 88.3 & 83.0 \\
DistilBERT & 67M & 6700M & 84.8 & 92.0 & 83.8 & 81.7 & 89.2 & 89.7 & 86.9 \\
TinyBERT-6L & 67M & 6700M & 85.3 & 89.0 & 86.2 & 85.7 & 89.3 & 90.0 & 87.6 \\
BERT-of-Theseus & 67M & 6700M & 84.4 & 90.7 & 82.4 & 85.0 & 89.4 & 92.1 & 87.3 \\
\textbf{ElasticBERT}\textsubscript{BASE}-6L & 67M & 6700M & \textbf{87.0} & 92.7 & \textbf{87.3} & \textbf{86.9} & 90.1 & \textbf{92.5} & \textbf{89.4} \\
\midrule
\multicolumn{10}{c}{\textit{LARGE Models}} \\
\midrule
BERT\textsubscript{LARGE} & 335M & 47214M & 87.9 & 94.0 & 85.9 & 86.7 & 90.8 & 93.9 & 89.9 \\  
ALBERT\textsubscript{LARGE} & 18M & 48876M & 87.7 & 93.8 & 88.1 & 89.3 & 90.2 & 93.6 & 90.5 \\
RoBERTa\textsubscript{LARGE} & 355M & 46042M & \textbf{90.5} & \textbf{95.7} & \textbf{89.9} & 90.5 & \textbf{91.6} & \textbf{95.8} & \textbf{92.3} \\
LayerDrop\textsubscript{LARGE} & 355M & 46042M & 90.4 & 95.3 & 89.5 & \textbf{91.0} & 91.4 & 95.2 & 92.1 \\
\textbf{ElasticBERT}\textsubscript{LARGE} & 335M & 47214M & 89.8 & 95.0 & 89.8 & 90.9 & 91.4 & 95.7 & 92.1 \\
\midrule
BERT\textsubscript{LARGE}-6L & 108M & 11922M & 80.4 & 89.6 & 74.3 & 70.5 & 87.4 & 84.4 & 81.1 \\  
ALBERT\textsubscript{LARGE}-6L & 18M & 12397M & 84.5 & 92.0 & 84.7 & 85.1 & 89.4 & 90.8 & 87.8 \\
RoBERTa\textsubscript{LARGE}-6L & 129M & 11664M & 83.5 & 91.7 & 77.9 & 72.7 & 88.6 & 84.7 & 83.2 \\
LayerDrop\textsubscript{LARGE}-6L & 129M & 11664M & 85.4 & 92.5 & 77.3 & 75.9 & 88.8 & 84.1 & 84.0 \\
\textbf{ElasticBERT}\textsubscript{LARGE}-6L & 108M & 11922M & \textbf{86.8} & \textbf{92.9} & \textbf{86.2} & \textbf{86.3} & \textbf{89.8} & \textbf{92.4} & \textbf{89.1} \\
\bottomrule
\end{tabular}
}
\caption{ElasticBERT and static baseline performance on ELUE task test sets. We report the mean of Accuracy and F1 for MRPC, Pearson and Spearman correlation for STS-B and Accuracy for other tasks. The reported FLOPs is the average over all the datasets.\vspace{-0.5cm}}
\label{tab:elue_static}
\end{table*}

\begin{figure}[t]
    \centering
    \begin{subfigure}{\linewidth}
    \centering
    \begin{tikzpicture} [scale=0.8]
    \begin{axis}[
        enlargelimits=0.25,
        legend style={at={(0.5,-0.35)},
          anchor=north,legend columns=-1},
        ylabel={Average Perf},
        symbolic x coords={3L,4L,6L},
        xtick=data,
    	ybar=3pt,
    	bar width=7pt,
        font=\small,
        grid=major,
        width=\linewidth,
        height=.5\linewidth,
        ]
        \addplot 
        coordinates {(3L,81.8) (4L,84.1) (6L,86.5)};
    
        \addplot 
    	coordinates {(3L,84.6) (4L,85.8) (6L,87.9)};
    
        \addplot 
    	coordinates {(3L,81.4) (4L,85.6) (6L,88.7)};
    	
        \addplot 
    	coordinates {(3L,83.4) (4L,85.7) (6L,88.6)};
    
        \addplot 
    	coordinates {(3L,87.3) (4L,88.2) (6L,89.4)};
    
    \legend{BERT, ALBERT, RoBERTa, LayerDrop, ElasticBERT}
    \end{axis}
    \end{tikzpicture}
    \caption{BASE models.}
    \end{subfigure}
    \\
    \begin{subfigure}{\linewidth}
    \centering
        \begin{tikzpicture} [scale=0.8]
        \begin{axis}[
            enlargelimits=0.25,
            legend style={at={(0.5,-0.35)},
              anchor=north,legend columns=-1},
            ylabel={Average Perf},
            symbolic x coords={6L,8L,12L},
            xtick=data,
        	ybar=3pt,
        	bar width=7pt,
            grid=major,
            font=\small,
            width=\linewidth,
            height=.5\linewidth,
            ]
            \addplot 
            coordinates {(6L,81.1) (8L,83.5) (12L,85.8)};
        
            \addplot 
        	coordinates {(6L,87.9) (8L,89.4) (12L,90.0)};
        
            \addplot 
        	coordinates {(6L,83.2) (8L,85.4) (12L,90.9)};

            \addplot 
        	coordinates {(6L,84.0) (8L,87.0) (12L,90.9)};
        
            \addplot 
        	coordinates {(6L,89.1) (8L,89.7) (12L,90.9)};	
        
        \legend{BERT, ALBERT, RoBERTa, LayerDrop, ElasticBERT}
        \end{axis}
        \end{tikzpicture}
        \caption{LARGE models.}
    \end{subfigure}
    \caption{Comparison of the average performance on ELUE test sets between ElasticBERT and baselines.\vspace{-0.5cm}}
    \label{fig:elue_layer}
\end{figure}
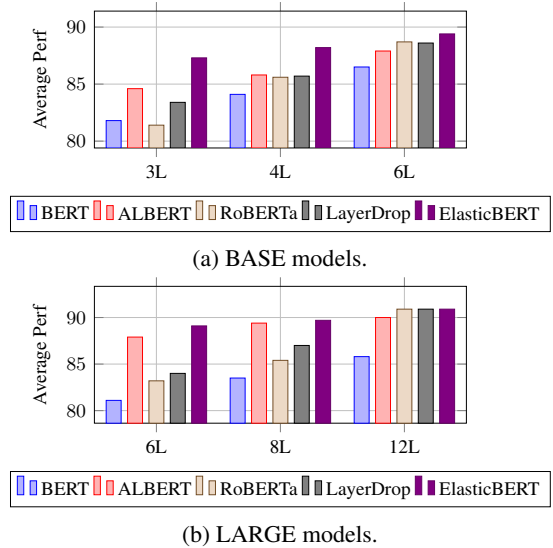

\paragraph{Baselines}
We compare ElasticBERT with three types of baselines: \textbf{(1)} Directly fine-tuning pre-trained models and their first $n$ layers. We choose BERT~\cite{Devlin2019BERT}, ALBERT~\cite{Lan2020ALBERT}, RoBERTa~\cite{Liu2019roberta} and LayerDrop~\cite{Fan2020LayerDrop} as our baselines. For the use of the first $n$ layers, we simply add a linear classifier on top of the truncated model. \textbf{(2)} Compressed models. We choose two distilled models, DistilBERT~\cite{Sanh2019DistilBERT} and TinyBERT~\cite{Jiao2020TinyBERT}, one pruned model, HeadPrune~\cite{michel19headprune}, and one model obtained by using \textit{module replacing},  BERT-of-Theseus~\cite{Xu2020BERTTheseus} as our baseline models. \textbf{(3)} Dynamic early exiting models. To verify the effectiveness of ElasticBERT as a strong backbone of dynamic early exiting methods, we also compare ElasticBERT\textsubscript{entropy} and ElasticBERT\textsubscript{patience} which have the same early exiting strategy as DeeBERT~\cite{Xin2020DeeBERT} and PABEE~\cite{Zhou2020PABEE} with four representative early exiting models: DeeBERT~\cite{Xin2020DeeBERT}, FastBERT~\cite{liu20fastbert}, PABEE~\cite{Zhou2020PABEE}, and CascadeBERT~\cite{Li2021cascadebert}.

\begin{figure*}[t!]
    \centering
    \begin{subfigure}{0.32\linewidth}
    \centering
    \includegraphics[width=\linewidth]{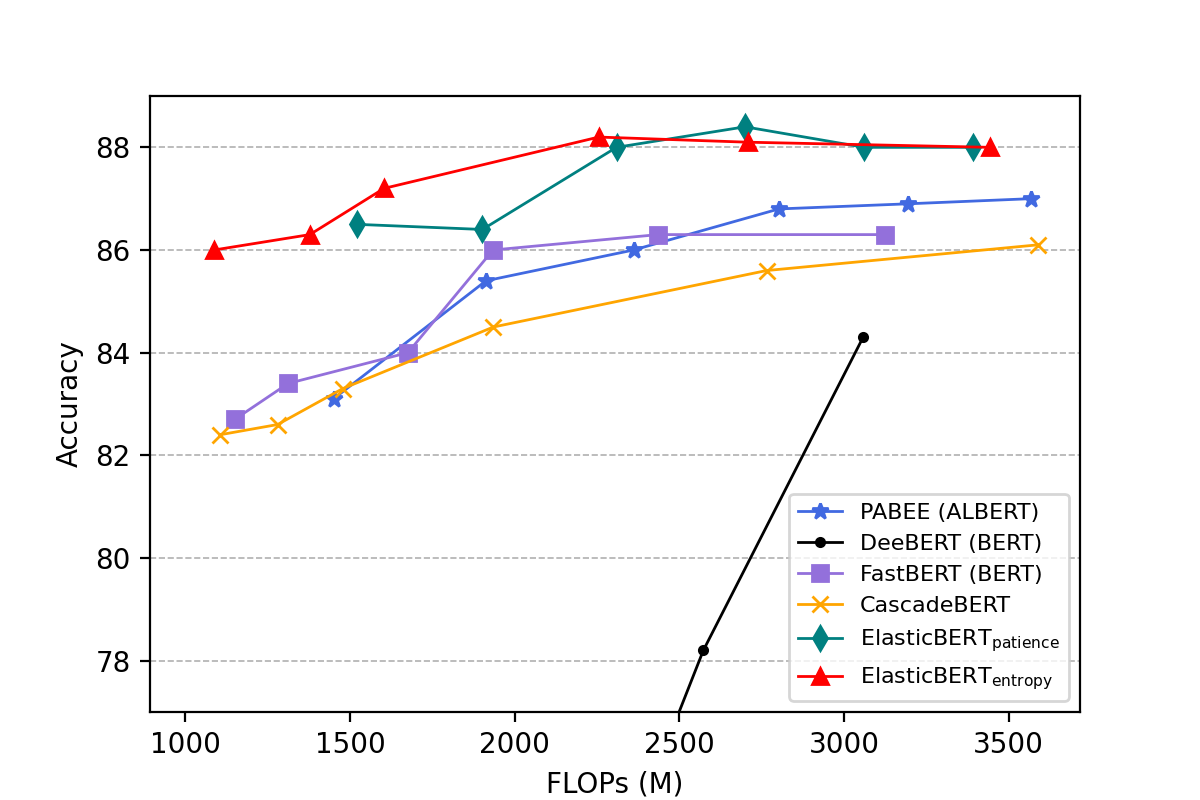}
    \caption{SST-2}
    \end{subfigure}
    \begin{subfigure}{0.32\linewidth}
    \centering
    \includegraphics[width=\linewidth]{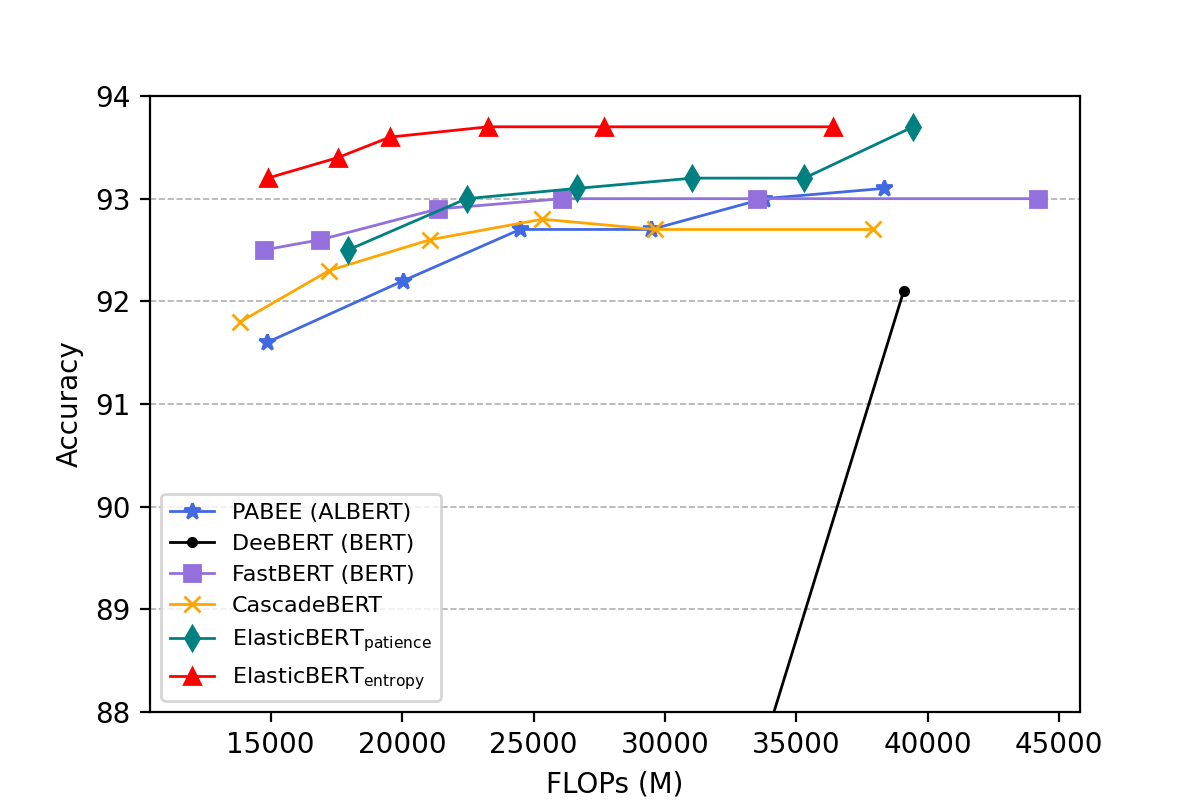}
    \caption{IMDb}
    \end{subfigure}
    \begin{subfigure}{0.32\linewidth}
    \centering
    \includegraphics[width=\linewidth]{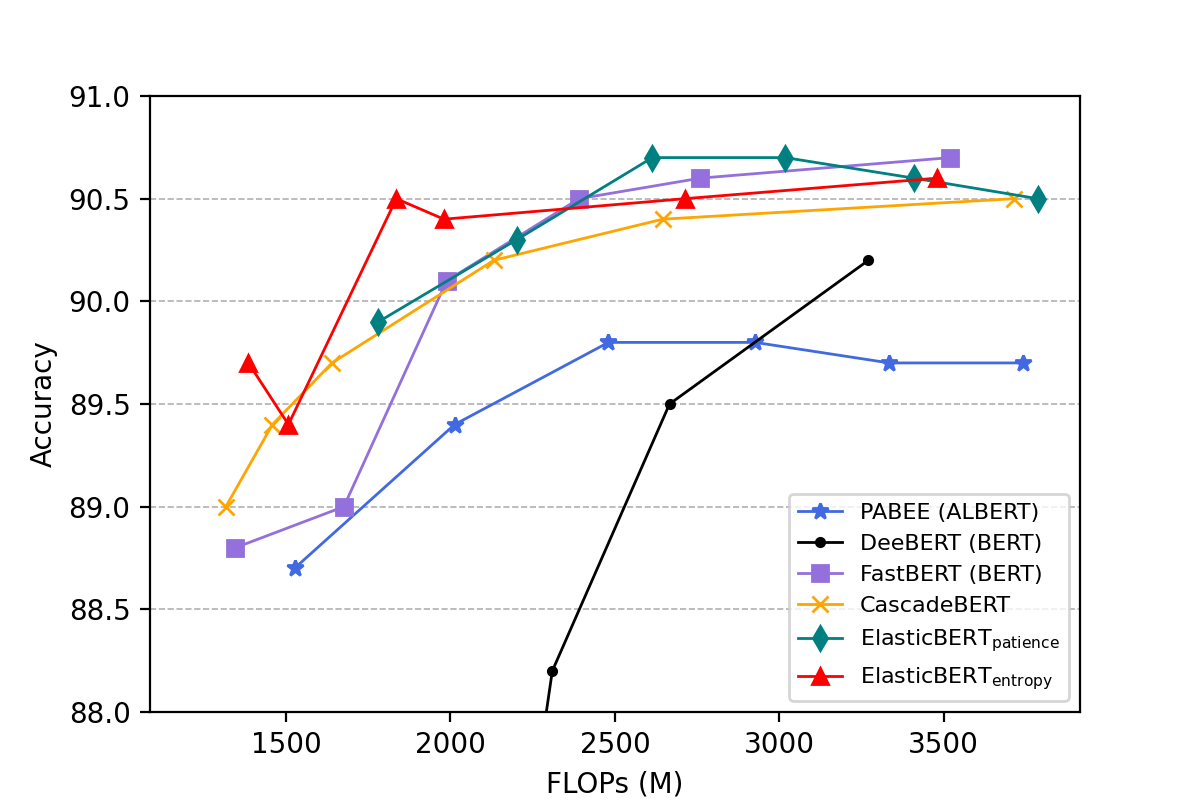}
    \caption{SNLI}
    \end{subfigure}
    \\
    \begin{subfigure}{0.32\linewidth}
    \centering
    \includegraphics[width=\linewidth]{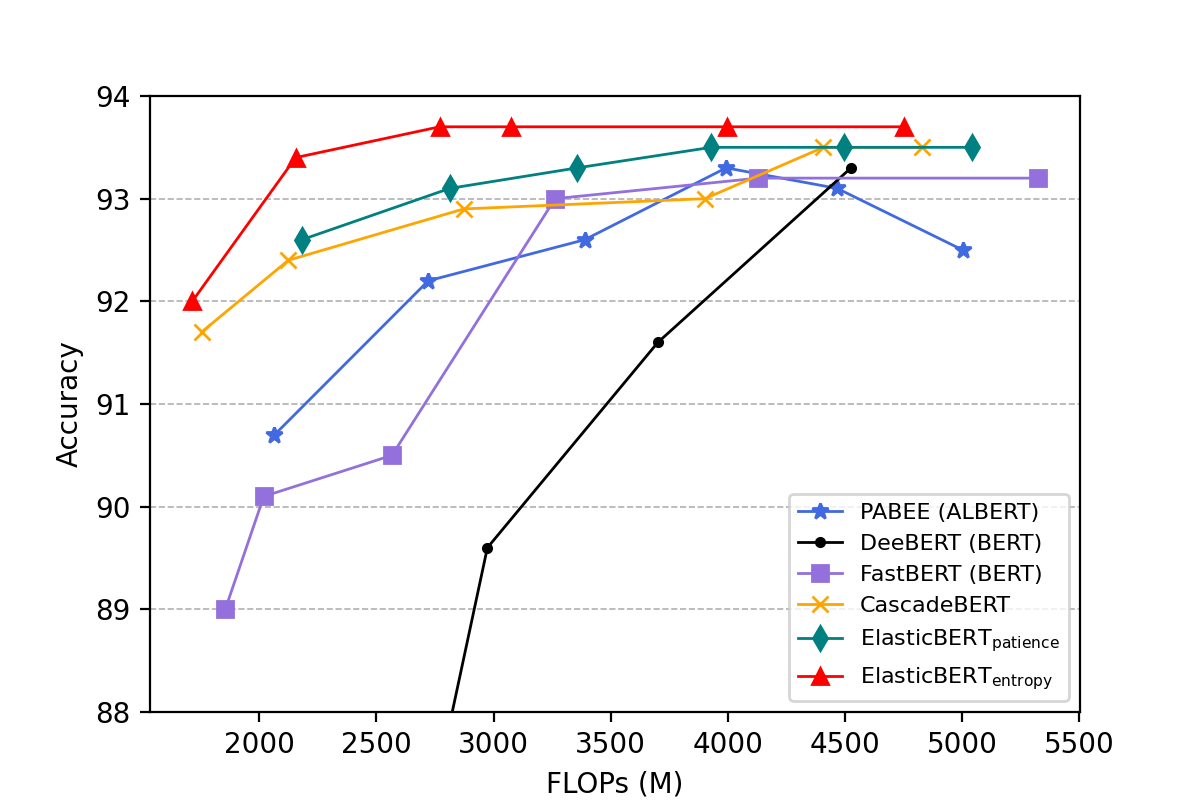}
    \caption{SciTail}
    \end{subfigure}
    \begin{subfigure}{0.32\linewidth}
    \centering
    \includegraphics[width=\linewidth]{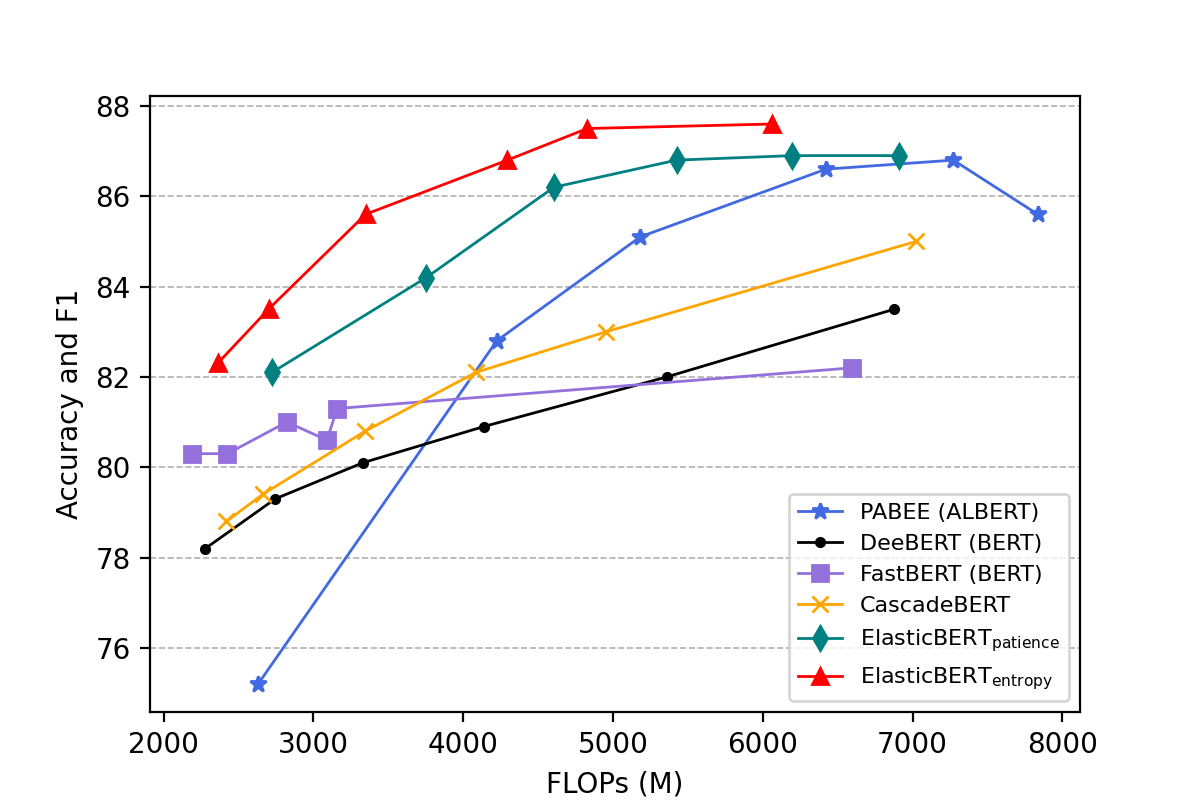}
    \caption{MRPC}
    \end{subfigure}
    \begin{subfigure}{0.32\linewidth}
    \centering
    \includegraphics[width=\linewidth]{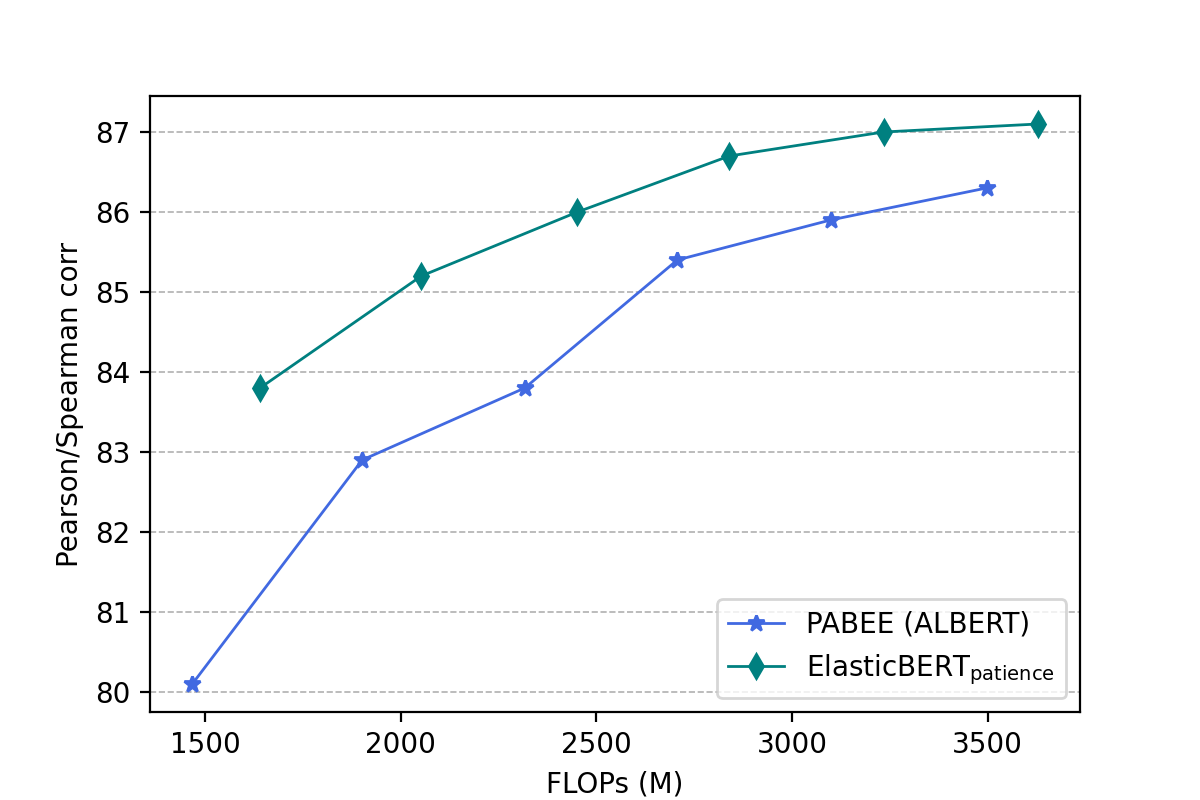}
    \caption{STS-B}
    \end{subfigure}
    \caption{Performance-FLOPs trade-offs on ELUE task test sets. Because STS-B is a regression task, for which the entropy-based methods are not applicable, we only evaluate patience-based methods, i.e., PABEE and ElasticBERT\textsubscript{patience}. }
    \label{fig:elue_dynamic}
\end{figure*}

\subsection{Evaluating ElasticBERT on ELUE}
ElasticBERT and our baselines are evaluated on ELUE tasks. For the BASE version of ElasticBERT, BERT, ALBERT, RoBERTa and LayerDrop, we evaluate the first 3/4/6/12 layers. For the LARGE version of the models, we evaluate the first 6/8/12/24 layers. For dynamic methods, we fine-tune ElasticBERT along with the injected internal classifiers using the gradient equilibrium (GE) strategy~\cite{Li2019Improve}, and adopt two different early exiting strategies: entropy-based strategy~\cite{Xin2020DeeBERT} and patience-based strategy~\cite{Zhou2020PABEE}.

\begin{table*}[t!]
\centering
\small
\resizebox{.85\linewidth}{!}{
\begin{tabular}{lccccccc}
\toprule
                Model     & \textbf{SST-2} & \textbf{IMDb}  & \textbf{MRPC} & \textbf{STS-B} & \textbf{SNLI}  & \textbf{SciTail} & \textbf{Average} \\
\midrule
                     
\textbf{ElasticBERT}\textsubscript{BASE}     & 0.00   & 0.00   & 0.00  & 0.00  & 0.00   & 0.00    & 0.00     \\
\midrule
\multicolumn{8}{c}{\textit{Static Models}} \\
\midrule
BERT\textsubscript{BASE}                 & -4.55  & -2.15  & -5.88 & -4.75 & -1.50  & -3.35   & -3.70    \\
ALBERT\textsubscript{BASE}              & -2.41  & -1.08  & -2.34 & -2.81 & -1.55  & -1.50   & \textbf{-1.95}    \\
RoBERTa\textsubscript{BASE}             & \textbf{-0.89}  & \textbf{-0.11}  & -2.95 & -5.38 & \textbf{-0.66}  & -3.32   & -2.22    \\
LayerDrop\textsubscript{BASE}             & -1.17  & -0.13 & \textbf{-2.17} & -2.98 & -1.36  & -4.14   & -1.99    \\
HeadPrune-BERT\textsubscript{BASE} & -3.81  & -8.61 & -9.73 & -11.9 & -2.89  & -4.18   & -6.85    \\
DistilBERT           & -2.20  & -0.70 & -3.50 & -5.20 & -0.90  & -2.80   & -2.55    \\
TinyBERT-6L          & -1.70  & -3.70  & -2.60 & -1.90 & -0.80  & -2.50   & -2.20    \\
BERT-of-Theseus      & -4.21  & -2.61  & -5.13 & \textbf{-1.67} & -1.29  & \textbf{-0.38}   & -2.55    \\
\midrule
\multicolumn{8}{c}{\textit{Dynamic Models}} \\
\midrule
PABEE                & -1.33  & -0.23  & -2.93 & -2.13 & -0.85  & -0.43   & -1.50    \\
DeeBERT         & -12.1 & -14.0 & -4.88 & -     & -8.35  & -6.19   & -        \\
FastBERT      & -1.51  & 0.16  & -3.70 & -     & -0.22 & -1.23   & -       \\
CascadeBERT      & -2.13  & -0.12  & -4.05 & -     & -0.23 & 0.14   & -       \\
\textbf{ElasticBERT}\textsubscript{patience} & 0.40   & 0.20   & -1.00 & \textbf{-0.44} & \textbf{0.03}   & 0.36    & \textbf{-0.08}    \\
\textbf{ElasticBERT}\textsubscript{entropy}  & \textbf{0.97}   & \textbf{1.02}   & \textbf{-0.14} & -     & 0.02   & \textbf{0.64}    & -   \\
\bottomrule
\end{tabular}
}
\caption{ELUE scores calculated using Eq. (\ref{eq:elue_score}) for static and dynamic baseline models. '-' denotes that the dataset/metric is not applicable to the model.\vspace{-0.5cm}} 
\label{tab:elue scores}
\end{table*}

\paragraph{Results of Static Models}
The performance of ElasticBERT and our baseline models on ELUE task test sets is shown in Table~\ref{tab:elue_static}, where we find that ElasticBERT\textsubscript{BASE} and ElasticBERT\textsubscript{LARGE} outperform BERT and ALBERT with the same number of layers, but are slightly weaker than RoBERTa\textsubscript{BASE} and RoBERTa\textsubscript{LARGE}. Besides, we find that the superiority of ElasticBERT over its baselines can be significant with fewer layers (See Figure~\ref{fig:elue_layer} for the results of 3/4 (6/8) layers of the BASE (LARGE) models).

\paragraph{Results of Dynamic Models}
We compare ElasticBERT\textsubscript{entropy} and ElasticBERT\textsubscript{patience} with four dynamic models: DeeBERT~\cite{Xin2020DeeBERT}, FastBERT~\cite{liu20fastbert}, PABEE~\cite{Zhou2020PABEE}, and CascadeBERT~\cite{Li2021cascadebert}. The performance-FLOPs trade-off of the dynamic models on ELUE task test sets are shown in Figure~\ref{fig:elue_dynamic}, which demonstrates that ElasticBERT can achieve better performance-FLOPs trade-off.

\paragraph{Evaluating ELUE Scores}
According to Eq. (\ref{eq:elue_score}), we also evaluate the ELUE scores of these baselines. As shown in Table \ref{tab:elue scores}, the ELUE score of ElasticBERT\textsubscript{BASE} is natural to be zero on all tasks. Among the other baselines, we find that ElasticBERT\textsubscript{patience} achieves the best ELUE score, while HeadPrune achieves the worst ELUE score. In addition, we find that dynamic models perform better than static models on average.

\section{Conclusion and Future Work}
In this work, we present ELUE, which is a public benchmark and platform for efficient models, and ElasticBERT, which is a strong baseline (backbone) for efficient static (dynamic) models. Both of the two main contributions are aimed to build the Pareto frontier for NLU tasks, such that the position of existing work can be clearly recognized, and future work can be easily and fairly measured.

Our future work is mainly in four aspects: (1) Including more baselines in ELUE, (2) Supporting the evaluation for more frameworks such as TensorFlow~\cite{Abadi2016TensorFlow}, (3) Supporting diagnostics for submissions, (4) Supporting the evaluation of more different types of tasks.

\section*{Acknowledgment}
This work was supported by the National Key Research and Development Program of China (No.2020AAA0106702) and National Natural Science Foundation of China (No.62022027). 

\section*{Ethical Considerations}
The proposed ELUE benchmark aims to standardize efficiency measurement of NLP models. The collected datasets are widely used in previous work and, to our knowledge, do not have any attached privacy or ethical issues. Our proposed ElasticBERT is a pre-trained model to reduce computation cost and carbon emission. The pre-training data is public resources adopted in previous work, and therefore would not introduce new ethical concerns.



\bibliography{custom}

\begin{thebibliography}{64}
\expandafter\ifx\csname natexlab\endcsname\relax\def\natexlab#1{#1}\fi

\bibitem[{Abadi et~al.(2016)Abadi, Barham, Chen, Chen, Davis, Dean, Devin,
  Ghemawat, Irving, Isard, Kudlur, Levenberg, Monga, Moore, Murray, Steiner,
  Tucker, Vasudevan, Warden, Wicke, Yu, and Zheng}]{Abadi2016TensorFlow}
Mart{\'{\i}}n Abadi, Paul Barham, Jianmin Chen, Zhifeng Chen, Andy Davis,
  Jeffrey Dean, Matthieu Devin, Sanjay Ghemawat, Geoffrey Irving, Michael
  Isard, Manjunath Kudlur, Josh Levenberg, Rajat Monga, Sherry Moore,
  Derek~Gordon Murray, Benoit Steiner, Paul~A. Tucker, Vijay Vasudevan, Pete
  Warden, Martin Wicke, Yuan Yu, and Xiaoqiang Zheng. 2016.
\newblock \href
  {https://www.usenix.org/conference/osdi16/technical-sessions/presentation/abadi}
  {Tensorflow: {A} system for large-scale machine learning}.
\newblock In \emph{12th {USENIX} Symposium on Operating Systems Design and
  Implementation, {OSDI} 2016, Savannah, GA, USA, November 2-4, 2016}, pages
  265--283. {USENIX} Association.

\bibitem[{Bengio et~al.(2013)Bengio, L{\'{e}}onard, and
  Courville}]{Bengio2013Estimating}
Yoshua Bengio, Nicholas L{\'{e}}onard, and Aaron~C. Courville. 2013.
\newblock \href {http://arxiv.org/abs/1308.3432} {Estimating or propagating
  gradients through stochastic neurons for conditional computation}.
\newblock \emph{CoRR}, abs/1308.3432.

\bibitem[{Bowman et~al.(2015)Bowman, Angeli, Potts, and
  Manning}]{Bowman2015SNLI}
Samuel~R. Bowman, Gabor Angeli, Christopher Potts, and Christopher~D. Manning.
  2015.
\newblock \href {https://doi.org/10.18653/v1/d15-1075} {A large annotated
  corpus for learning natural language inference}.
\newblock In \emph{Proceedings of the 2015 Conference on Empirical Methods in
  Natural Language Processing, {EMNLP} 2015, Lisbon, Portugal, September 17-21,
  2015}, pages 632--642. The Association for Computational Linguistics.

\bibitem[{Brown et~al.(2020)Brown, Mann, Ryder, Subbiah, Kaplan, Dhariwal,
  Neelakantan, Shyam, Sastry, Askell, Agarwal, Herbert{-}Voss, Krueger,
  Henighan, Child, Ramesh, Ziegler, Wu, Winter, Hesse, Chen, Sigler, Litwin,
  Gray, Chess, Clark, Berner, McCandlish, Radford, Sutskever, and
  Amodei}]{Brown2020GPT3}
Tom~B. Brown, Benjamin Mann, Nick Ryder, Melanie Subbiah, Jared Kaplan,
  Prafulla Dhariwal, Arvind Neelakantan, Pranav Shyam, Girish Sastry, Amanda
  Askell, Sandhini Agarwal, Ariel Herbert{-}Voss, Gretchen Krueger, Tom
  Henighan, Rewon Child, Aditya Ramesh, Daniel~M. Ziegler, Jeffrey Wu, Clemens
  Winter, Christopher Hesse, Mark Chen, Eric Sigler, Mateusz Litwin, Scott
  Gray, Benjamin Chess, Jack Clark, Christopher Berner, Sam McCandlish, Alec
  Radford, Ilya Sutskever, and Dario Amodei. 2020.
\newblock \href
  {https://proceedings.neurips.cc/paper/2020/hash/1457c0d6bfcb4967418bfb8ac142f64a-Abstract.html}
  {Language models are few-shot learners}.
\newblock In \emph{Advances in Neural Information Processing Systems 33: Annual
  Conference on Neural Information Processing Systems 2020, NeurIPS 2020,
  December 6-12, 2020, virtual}.

\bibitem[{Cer et~al.(2017)Cer, Diab, Agirre, Lopez{-}Gazpio, and
  Specia}]{Cer2017SemEval2017}
Daniel~M. Cer, Mona~T. Diab, Eneko Agirre, I{\~{n}}igo Lopez{-}Gazpio, and
  Lucia Specia. 2017.
\newblock \href {http://arxiv.org/abs/1708.00055} {Semeval-2017 task 1:
  Semantic textual similarity - multilingual and cross-lingual focused
  evaluation}.
\newblock \emph{CoRR}, abs/1708.00055.

\bibitem[{Conneau and Kiela(2018)}]{Conneau2018SentEval}
Alexis Conneau and Douwe Kiela. 2018.
\newblock \href
  {http://www.lrec-conf.org/proceedings/lrec2018/summaries/757.html} {Senteval:
  An evaluation toolkit for universal sentence representations}.
\newblock In \emph{Proceedings of the Eleventh International Conference on
  Language Resources and Evaluation, {LREC} 2018, Miyazaki, Japan, May 7-12,
  2018}. European Language Resources Association {(ELRA)}.

\bibitem[{Davis and Arel(2014)}]{Davis2013Lowrank}
Andrew~S. Davis and Itamar Arel. 2014.
\newblock \href {http://arxiv.org/abs/1312.4461} {Low-rank approximations for
  conditional feedforward computation in deep neural networks}.
\newblock In \emph{2nd International Conference on Learning Representations,
  {ICLR} 2014, Banff, AB, Canada, April 14-16, 2014, Workshop Track
  Proceedings}.

\bibitem[{Dehghani et~al.(2019)Dehghani, Gouws, Vinyals, Uszkoreit, and
  Kaiser}]{Dehghani2019Universal}
Mostafa Dehghani, Stephan Gouws, Oriol Vinyals, Jakob Uszkoreit, and Lukasz
  Kaiser. 2019.
\newblock \href {https://openreview.net/forum?id=HyzdRiR9Y7} {Universal
  transformers}.
\newblock In \emph{7th International Conference on Learning Representations,
  {ICLR} 2019, New Orleans, LA, USA, May 6-9, 2019}. OpenReview.net.

\bibitem[{Devlin et~al.(2019)Devlin, Chang, Lee, and
  Toutanova}]{Devlin2019BERT}
Jacob Devlin, Ming{-}Wei Chang, Kenton Lee, and Kristina Toutanova. 2019.
\newblock \href {https://doi.org/10.18653/v1/n19-1423} {{BERT:} pre-training of
  deep bidirectional transformers for language understanding}.
\newblock In \emph{Proceedings of the 2019 Conference of the North American
  Chapter of the Association for Computational Linguistics: Human Language
  Technologies, {NAACL-HLT} 2019, Minneapolis, MN, USA, June 2-7, 2019, Volume
  1 (Long and Short Papers)}, pages 4171--4186. Association for Computational
  Linguistics.

\bibitem[{Dolan and Brockett(2005)}]{Dolan2005MRPC}
William~B. Dolan and Chris Brockett. 2005.
\newblock \href {https://aclanthology.org/I05-5002/} {Automatically
  constructing a corpus of sentential paraphrases}.
\newblock In \emph{Proceedings of the Third International Workshop on
  Paraphrasing, IWP@IJCNLP 2005, Jeju Island, Korea, October 2005, 2005}. Asian
  Federation of Natural Language Processing.

\bibitem[{Elbayad et~al.(2020)Elbayad, Gu, Grave, and Auli}]{Elbayad2020Depth}
Maha Elbayad, Jiatao Gu, Edouard Grave, and Michael Auli. 2020.
\newblock \href {https://openreview.net/forum?id=SJg7KhVKPH} {Depth-adaptive
  transformer}.
\newblock In \emph{8th International Conference on Learning Representations,
  {ICLR} 2020, Addis Ababa, Ethiopia, April 26-30, 2020}. OpenReview.net.

\bibitem[{Fan et~al.(2020)Fan, Grave, and Joulin}]{Fan2020LayerDrop}
Angela Fan, Edouard Grave, and Armand Joulin. 2020.
\newblock \href {https://openreview.net/forum?id=SylO2yStDr} {Reducing
  transformer depth on demand with structured dropout}.
\newblock In \emph{8th International Conference on Learning Representations,
  {ICLR} 2020, Addis Ababa, Ethiopia, April 26-30, 2020}. OpenReview.net.

\bibitem[{Gehrmann et~al.(2021)Gehrmann, Adewumi, Aggarwal, Ammanamanchi,
  Anuoluwapo, Bosselut, Chandu, Clinciu, Das, Dhole, Du, Durmus, Dusek, Emezue,
  Gangal, Garbacea, Hashimoto, Hou, Jernite, Jhamtani, Ji, Jolly, Kumar,
  Ladhak, Madaan, Maddela, Mahajan, Mahamood, Majumder, Martins,
  McMillan{-}Major, Mille, van Miltenburg, Nadeem, Narayan, Nikolaev,
  Niyongabo, Osei, Parikh, Perez{-}Beltrachini, Rao, Raunak, Rodriguez,
  Santhanam, Sedoc, Sellam, Shaikh, Shimorina, Cabezudo, Strobelt, Subramani,
  Xu, Yang, Yerukola, and Zhou}]{gem2021Gehrmann}
Sebastian Gehrmann, Tosin~P. Adewumi, Karmanya Aggarwal, Pawan~Sasanka
  Ammanamanchi, Aremu Anuoluwapo, Antoine Bosselut, Khyathi~Raghavi Chandu,
  Miruna{-}Adriana Clinciu, Dipanjan Das, Kaustubh~D. Dhole, Wanyu Du, Esin
  Durmus, Ondrej Dusek, Chris Emezue, Varun Gangal, Cristina Garbacea,
  Tatsunori Hashimoto, Yufang Hou, Yacine Jernite, Harsh Jhamtani, Yangfeng Ji,
  Shailza Jolly, Dhruv Kumar, Faisal Ladhak, Aman Madaan, Mounica Maddela,
  Khyati Mahajan, Saad Mahamood, Bodhisattwa~Prasad Majumder, Pedro~Henrique
  Martins, Angelina McMillan{-}Major, Simon Mille, Emiel van Miltenburg, Moin
  Nadeem, Shashi Narayan, Vitaly Nikolaev, Rubungo~Andre Niyongabo, Salomey
  Osei, Ankur~P. Parikh, Laura Perez{-}Beltrachini, Niranjan~Ramesh Rao, Vikas
  Raunak, Juan~Diego Rodriguez, Sashank Santhanam, Jo{\~{a}}o Sedoc, Thibault
  Sellam, Samira Shaikh, Anastasia Shimorina, Marco Antonio~Sobrevilla
  Cabezudo, Hendrik Strobelt, Nishant Subramani, Wei Xu, Diyi Yang, Akhila
  Yerukola, and Jiawei Zhou. 2021.
\newblock \href {http://arxiv.org/abs/2102.01672} {The {GEM} benchmark: Natural
  language generation, its evaluation and metrics}.
\newblock \emph{CoRR}, abs/2102.01672.

\bibitem[{Gokaslan and Cohen(2019)}]{Gokaslan2019OpenWeb}
Aaron Gokaslan and Vanya Cohen. 2019.
\newblock \href {http://Skylion007.github.io/OpenWebTextCorpus} {Openwebtext
  corpus}.

\bibitem[{Gordon et~al.(2020)Gordon, Duh, and Andrews}]{Gordon2020Compressing}
Mitchell~A. Gordon, Kevin Duh, and Nicholas Andrews. 2020.
\newblock \href {https://doi.org/10.18653/v1/2020.repl4nlp-1.18} {Compressing
  {BERT:} studying the effects of weight pruning on transfer learning}.
\newblock In \emph{Proceedings of the 5th Workshop on Representation Learning
  for NLP, RepL4NLP@ACL 2020, Online, July 9, 2020}, pages 143--155.
  Association for Computational Linguistics.

\bibitem[{Graves(2016)}]{Graves2016Adaptive}
Alex Graves. 2016.
\newblock \href {http://arxiv.org/abs/1603.08983} {Adaptive computation time
  for recurrent neural networks}.
\newblock \emph{CoRR}, abs/1603.08983.

\bibitem[{Jiao et~al.(2020)Jiao, Yin, Shang, Jiang, Chen, Li, Wang, and
  Liu}]{Jiao2020TinyBERT}
Xiaoqi Jiao, Yichun Yin, Lifeng Shang, Xin Jiang, Xiao Chen, Linlin Li, Fang
  Wang, and Qun Liu. 2020.
\newblock \href {https://doi.org/10.18653/v1/2020.findings-emnlp.372}
  {Tinybert: Distilling {BERT} for natural language understanding}.
\newblock In \emph{Proceedings of {EMNLP} 2020}, pages 4163--4174. Association
  for Computational Linguistics.

\bibitem[{Khot et~al.(2018)Khot, Sabharwal, and Clark}]{Khot2018SciTail}
Tushar Khot, Ashish Sabharwal, and Peter Clark. 2018.
\newblock \href
  {https://www.aaai.org/ocs/index.php/AAAI/AAAI18/paper/view/17368} {Scitail:
  {A} textual entailment dataset from science question answering}.
\newblock In \emph{Proceedings of the Thirty-Second {AAAI} Conference on
  Artificial Intelligence, (AAAI-18), the 30th innovative Applications of
  Artificial Intelligence (IAAI-18), and the 8th {AAAI} Symposium on
  Educational Advances in Artificial Intelligence (EAAI-18), New Orleans,
  Louisiana, USA, February 2-7, 2018}, pages 5189--5197. {AAAI} Press.

\bibitem[{Kiela et~al.(2021)Kiela, Bartolo, Nie, Kaushik, Geiger, Wu, Vidgen,
  Prasad, Singh, Ringshia, Ma, Thrush, Riedel, Waseem, Stenetorp, Jia, Bansal,
  Potts, and Williams}]{Kiela2021Dynabench}
Douwe Kiela, Max Bartolo, Yixin Nie, Divyansh Kaushik, Atticus Geiger,
  Zhengxuan Wu, Bertie Vidgen, Grusha Prasad, Amanpreet Singh, Pratik Ringshia,
  Zhiyi Ma, Tristan Thrush, Sebastian Riedel, Zeerak Waseem, Pontus Stenetorp,
  Robin Jia, Mohit Bansal, Christopher Potts, and Adina Williams. 2021.
\newblock \href {https://doi.org/10.18653/v1/2021.naacl-main.324} {Dynabench:
  Rethinking benchmarking in {NLP}}.
\newblock In \emph{Proceedings of the 2021 Conference of the North American
  Chapter of the Association for Computational Linguistics: Human Language
  Technologies, {NAACL-HLT} 2021, Online, June 6-11, 2021}, pages 4110--4124.
  Association for Computational Linguistics.

\bibitem[{Kingma and Ba(2015)}]{kingma2015adam}
Diederik~P. Kingma and Jimmy Ba. 2015.
\newblock \href {http://arxiv.org/abs/1412.6980} {Adam: {A} method for
  stochastic optimization}.
\newblock In \emph{3rd International Conference on Learning Representations,
  {ICLR} 2015, San Diego, CA, USA, May 7-9, 2015, Conference Track
  Proceedings}.

\bibitem[{Lan et~al.(2020)Lan, Chen, Goodman, Gimpel, Sharma, and
  Soricut}]{Lan2020ALBERT}
Zhenzhong Lan, Mingda Chen, Sebastian Goodman, Kevin Gimpel, Piyush Sharma, and
  Radu Soricut. 2020.
\newblock \href {https://openreview.net/forum?id=H1eA7AEtvS} {{ALBERT:} {A}
  lite {BERT} for self-supervised learning of language representations}.
\newblock In \emph{8th International Conference on Learning Representations,
  {ICLR} 2020, Addis Ababa, Ethiopia, April 26-30, 2020}. OpenReview.net.

\bibitem[{Li et~al.(2019)Li, Zhang, Qi, Yang, and Huang}]{Li2019Improve}
Hao Li, Hong Zhang, Xiaojuan Qi, Ruigang Yang, and Gao Huang. 2019.
\newblock \href {https://doi.org/10.1109/ICCV.2019.00198} {Improved techniques
  for training adaptive deep networks}.
\newblock In \emph{2019 {IEEE/CVF} International Conference on Computer Vision,
  {ICCV} 2019, Seoul, Korea (South), October 27 - November 2, 2019}, pages
  1891--1900. {IEEE}.

\bibitem[{Li et~al.(2021{\natexlab{a}})Li, Lin, Chen, Ren, Li, Zhou, and
  Sun}]{Li2021cascadebert}
Lei Li, Yankai Lin, Deli Chen, Shuhuai Ren, Peng Li, Jie Zhou, and Xu~Sun.
  2021{\natexlab{a}}.
\newblock Cascadebert: Accelerating inference of pre-trained language models
  via calibrated complete models cascade.
\newblock In \emph{Findings of EMNLP}.

\bibitem[{Li et~al.(2021{\natexlab{b}})Li, Shao, Sun, Yan, Qiu, and
  Huang}]{Li2020Accelerating}
Xiaonan Li, Yunfan Shao, Tianxiang Sun, Hang Yan, Xipeng Qiu, and Xuanjing
  Huang. 2021{\natexlab{b}}.
\newblock \href {https://doi.org/10.18653/v1/2021.acl-long.16} {Accelerating
  {BERT} inference for sequence labeling via early-exit}.
\newblock In \emph{Proceedings of {ACL/IJCNLP} 2021}, pages 189--199.
  Association for Computational Linguistics.

\bibitem[{Liao et~al.(2021)Liao, Zhang, Ren, Su, Sun, and He}]{Liao2021Global}
Kaiyuan Liao, Yi~Zhang, Xuancheng Ren, Qi~Su, Xu~Sun, and Bin He. 2021.
\newblock \href {https://doi.org/10.18653/v1/2021.naacl-main.162} {A global
  past-future early exit method for accelerating inference of pre-trained
  language models}.
\newblock In \emph{Proceedings of the 2021 Conference of the North American
  Chapter of the Association for Computational Linguistics: Human Language
  Technologies, {NAACL-HLT} 2021, Online, June 6-11, 2021}, pages 2013--2023.
  Association for Computational Linguistics.

\bibitem[{Lin et~al.(2021)Lin, Wang, Liu, and Qiu}]{lin2021transformers}
Tianyang Lin, Yuxin Wang, Xiangyang Liu, and Xipeng Qiu. 2021.
\newblock \href {http://arxiv.org/abs/2106.04554} {A survey of transformers}.
\newblock \emph{CoRR}, abs/2106.04554.

\bibitem[{Liu et~al.(2020{\natexlab{a}})Liu, Zhou, Wang, Zhao, Deng, and
  Ju}]{Liu2020FastBERT}
Weijie Liu, Peng Zhou, Zhiruo Wang, Zhe Zhao, Haotang Deng, and Qi~Ju.
  2020{\natexlab{a}}.
\newblock \href {https://www.aclweb.org/anthology/2020.acl-main.537/}
  {Fastbert: a self-distilling {BERT} with adaptive inference time}.
\newblock In \emph{Proceedings of the 58th Annual Meeting of the Association
  for Computational Linguistics, {ACL} 2020, Online, July 5-10, 2020}, pages
  6035--6044. Association for Computational Linguistics.

\bibitem[{Liu et~al.(2020{\natexlab{b}})Liu, Zhou, Wang, Zhao, Deng, and
  Ju}]{liu20fastbert}
Weijie Liu, Peng Zhou, Zhiruo Wang, Zhe Zhao, Haotang Deng, and Qi~Ju.
  2020{\natexlab{b}}.
\newblock \href {https://doi.org/10.18653/v1/2020.acl-main.537} {Fastbert: a
  self-distilling {BERT} with adaptive inference time}.
\newblock In \emph{Proceedings of the 58th Annual Meeting of the Association
  for Computational Linguistics, {ACL} 2020, Online, July 5-10, 2020}, pages
  6035--6044. Association for Computational Linguistics.

\bibitem[{Liu et~al.(2019)Liu, Ott, Goyal, Du, Joshi, Chen, Levy, Lewis,
  Zettlemoyer, and Stoyanov}]{Liu2019roberta}
Yinhan Liu, Myle Ott, Naman Goyal, Jingfei Du, Mandar Joshi, Danqi Chen, Omer
  Levy, Mike Lewis, Luke Zettlemoyer, and Veselin Stoyanov. 2019.
\newblock \href {http://arxiv.org/abs/1907.11692} {Roberta: {A} robustly
  optimized {BERT} pretraining approach}.
\newblock \emph{CoRR}, abs/1907.11692.

\bibitem[{Loshchilov and Hutter(2019)}]{Ilya2019AdamW}
Ilya Loshchilov and Frank Hutter. 2019.
\newblock \href {https://openreview.net/forum?id=Bkg6RiCqY7} {Decoupled weight
  decay regularization}.
\newblock In \emph{7th International Conference on Learning Representations,
  {ICLR} 2019, New Orleans, LA, USA, May 6-9, 2019}. OpenReview.net.

\bibitem[{Maas et~al.(2011)Maas, Daly, Pham, Huang, Ng, and
  Potts}]{Maas2011IMDb}
Andrew~L. Maas, Raymond~E. Daly, Peter~T. Pham, Dan Huang, Andrew~Y. Ng, and
  Christopher Potts. 2011.
\newblock \href {https://aclanthology.org/P11-1015/} {Learning word vectors for
  sentiment analysis}.
\newblock In \emph{The 49th Annual Meeting of the Association for Computational
  Linguistics: Human Language Technologies, Proceedings of the Conference,
  19-24 June, 2011, Portland, Oregon, {USA}}, pages 142--150. The Association
  for Computer Linguistics.

\bibitem[{McCann et~al.(2018)McCann, Keskar, Xiong, and
  Socher}]{McCann2018DecaNLP}
Bryan McCann, Nitish~Shirish Keskar, Caiming Xiong, and Richard Socher. 2018.
\newblock \href {http://arxiv.org/abs/1806.08730} {The natural language
  decathlon: Multitask learning as question answering}.
\newblock \emph{CoRR}, abs/1806.08730.

\bibitem[{Michel et~al.(2019)Michel, Levy, and Neubig}]{michel19headprune}
Paul Michel, Omer Levy, and Graham Neubig. 2019.
\newblock \href
  {https://proceedings.neurips.cc/paper/2019/hash/2c601ad9d2ff9bc8b282670cdd54f69f-Abstract.html}
  {Are sixteen heads really better than one?}
\newblock In \emph{Advances in Neural Information Processing Systems 32: Annual
  Conference on Neural Information Processing Systems 2019, NeurIPS 2019,
  December 8-14, 2019, Vancouver, BC, Canada}, pages 14014--14024.

\bibitem[{Min et~al.(2020)Min, Boyd{-}Graber, Alberti, Chen, Choi, Collins,
  Guu, Hajishirzi, Lee, Palomaki, Raffel, Roberts, Kwiatkowski, Lewis, Wu,
  K{\"{u}}ttler, Liu, Minervini, Stenetorp, Riedel, Yang, Seo, Izacard,
  Petroni, Hosseini, Cao, Grave, Yamada, Shimaoka, Suzuki, Miyawaki, Sato,
  Takahashi, Suzuki, Fajcik, Docekal, Ondrej, Smrz, Cheng, Shen, Liu, He, Chen,
  Gao, Oguz, Chen, Karpukhin, Peshterliev, Okhonko, Schlichtkrull, Gupta,
  Mehdad, and Yih}]{Min2020EfficientQA}
Sewon Min, Jordan~L. Boyd{-}Graber, Chris Alberti, Danqi Chen, Eunsol Choi,
  Michael Collins, Kelvin Guu, Hannaneh Hajishirzi, Kenton Lee, Jennimaria
  Palomaki, Colin Raffel, Adam Roberts, Tom Kwiatkowski, Patrick S.~H. Lewis,
  Yuxiang Wu, Heinrich K{\"{u}}ttler, Linqing Liu, Pasquale Minervini, Pontus
  Stenetorp, Sebastian Riedel, Sohee Yang, Minjoon Seo, Gautier Izacard, Fabio
  Petroni, Lucas Hosseini, Nicola~De Cao, Edouard Grave, Ikuya Yamada, Sonse
  Shimaoka, Masatoshi Suzuki, Shumpei Miyawaki, Shun Sato, Ryo Takahashi, Jun
  Suzuki, Martin Fajcik, Martin Docekal, Karel Ondrej, Pavel Smrz, Hao Cheng,
  Yelong Shen, Xiaodong Liu, Pengcheng He, Weizhu Chen, Jianfeng Gao, Barlas
  Oguz, Xilun Chen, Vladimir Karpukhin, Stan Peshterliev, Dmytro Okhonko,
  Michael~Sejr Schlichtkrull, Sonal Gupta, Yashar Mehdad, and Wen{-}tau Yih.
  2020.
\newblock \href {http://proceedings.mlr.press/v133/min21a.html} {Neurips 2020
  efficientqa competition: Systems, analyses and lessons learned}.
\newblock In \emph{NeurIPS 2020 Competition and Demonstration Track, 6-12
  December 2020, Virtual Event / Vancouver, BC, Canada}, volume 133 of
  \emph{Proceedings of Machine Learning Research}, pages 86--111. {PMLR}.

\bibitem[{Paszke et~al.(2019)Paszke, Gross, Massa, Lerer, Bradbury, Chanan,
  Killeen, Lin, Gimelshein, Antiga, Desmaison, K{\"{o}}pf, Yang, DeVito,
  Raison, Tejani, Chilamkurthy, Steiner, Fang, Bai, and
  Chintala}]{Paszke2019Pytorch}
Adam Paszke, Sam Gross, Francisco Massa, Adam Lerer, James Bradbury, Gregory
  Chanan, Trevor Killeen, Zeming Lin, Natalia Gimelshein, Luca Antiga, Alban
  Desmaison, Andreas K{\"{o}}pf, Edward Yang, Zachary DeVito, Martin Raison,
  Alykhan Tejani, Sasank Chilamkurthy, Benoit Steiner, Lu~Fang, Junjie Bai, and
  Soumith Chintala. 2019.
\newblock \href
  {https://proceedings.neurips.cc/paper/2019/hash/bdbca288fee7f92f2bfa9f7012727740-Abstract.html}
  {Pytorch: An imperative style, high-performance deep learning library}.
\newblock In \emph{Advances in Neural Information Processing Systems 32: Annual
  Conference on Neural Information Processing Systems 2019, NeurIPS 2019,
  December 8-14, 2019, Vancouver, BC, Canada}, pages 8024--8035.

\bibitem[{Qiu et~al.(2020)Qiu, Sun, Xu, Shao, Dai, and Huang}]{Qiu2020survey}
Xipeng Qiu, Tianxiang Sun, Yige Xu, Yunfan Shao, Ning Dai, and Xuanjing Huang.
  2020.
\newblock \href {https://doi.org/https://doi.org/10.1007/s11431-020-1647-3}
  {Pre-trained models for natural language processing: {A} survey}.
\newblock \emph{SCIENCE CHINA Technological Sciences}.

\bibitem[{Raffel et~al.(2020)Raffel, Shazeer, Roberts, Lee, Narang, Matena,
  Zhou, Li, and Liu}]{Raffel2020T5}
Colin Raffel, Noam Shazeer, Adam Roberts, Katherine Lee, Sharan Narang, Michael
  Matena, Yanqi Zhou, Wei Li, and Peter~J. Liu. 2020.
\newblock \href {http://jmlr.org/papers/v21/20-074.html} {Exploring the limits
  of transfer learning with a unified text-to-text transformer}.
\newblock \emph{J. Mach. Learn. Res.}, 21:140:1--140:67.

\bibitem[{Rajpurkar et~al.(2016)Rajpurkar, Zhang, Lopyrev, and
  Liang}]{Rajpurkar2016SQuAD}
Pranav Rajpurkar, Jian Zhang, Konstantin Lopyrev, and Percy Liang. 2016.
\newblock \href {https://doi.org/10.18653/v1/d16-1264} {Squad: 100, 000+
  questions for machine comprehension of text}.
\newblock In \emph{Proceedings of the 2016 Conference on Empirical Methods in
  Natural Language Processing, {EMNLP} 2016, Austin, Texas, USA, November 1-4,
  2016}, pages 2383--2392. The Association for Computational Linguistics.

\bibitem[{Sanh et~al.(2019)Sanh, Debut, Chaumond, and
  Wolf}]{Sanh2019DistilBERT}
Victor Sanh, Lysandre Debut, Julien Chaumond, and Thomas Wolf. 2019.
\newblock \href {http://arxiv.org/abs/1910.01108} {Distilbert, a distilled
  version of {BERT:} smaller, faster, cheaper and lighter}.
\newblock \emph{CoRR}, abs/1910.01108.

\bibitem[{Schwartz et~al.(2020)Schwartz, Stanovsky, Swayamdipta, Dodge, and
  Smith}]{Schwartz2020Right}
Roy Schwartz, Gabriel Stanovsky, Swabha Swayamdipta, Jesse Dodge, and Noah~A.
  Smith. 2020.
\newblock \href {https://www.aclweb.org/anthology/2020.acl-main.593/} {The
  right tool for the job: Matching model and instance complexities}.
\newblock In \emph{Proceedings of the 58th Annual Meeting of the Association
  for Computational Linguistics, {ACL} 2020, Online, July 5-10, 2020}, pages
  6640--6651. Association for Computational Linguistics.

\bibitem[{Shen et~al.(2020)Shen, Dong, Ye, Ma, Yao, Gholami, Mahoney, and
  Keutzer}]{Shen2020QBERT}
Sheng Shen, Zhen Dong, Jiayu Ye, Linjian Ma, Zhewei Yao, Amir Gholami,
  Michael~W. Mahoney, and Kurt Keutzer. 2020.
\newblock \href {https://aaai.org/ojs/index.php/AAAI/article/view/6409}
  {{Q-BERT:} hessian based ultra low precision quantization of {BERT}}.
\newblock In \emph{The Thirty-Fourth {AAAI} Conference on Artificial
  Intelligence, {AAAI} 2020, New York, NY, USA, February 7-12, 2020}, pages
  8815--8821. {AAAI} Press.

\bibitem[{Shoeybi et~al.(2019)Shoeybi, Patwary, Puri, LeGresley, Casper, and
  Catanzaro}]{shoeybi2019megatron}
Mohammad Shoeybi, Mostofa Patwary, Raul Puri, Patrick LeGresley, Jared Casper,
  and Bryan Catanzaro. 2019.
\newblock \href {http://arxiv.org/abs/1909.08053} {Megatron-lm: Training
  multi-billion parameter language models using model parallelism}.
\newblock \emph{CoRR}, abs/1909.08053.

\bibitem[{Socher et~al.(2013)Socher, Perelygin, Wu, Chuang, Manning, Ng, and
  Potts}]{Socher2013SST}
Richard Socher, Alex Perelygin, Jean Wu, Jason Chuang, Christopher~D. Manning,
  Andrew~Y. Ng, and Christopher Potts. 2013.
\newblock \href {https://aclanthology.org/D13-1170/} {Recursive deep models for
  semantic compositionality over a sentiment treebank}.
\newblock In \emph{Proceedings of the 2013 Conference on Empirical Methods in
  Natural Language Processing, {EMNLP} 2013, 18-21 October 2013, Grand Hyatt
  Seattle, Seattle, Washington, USA, {A} meeting of SIGDAT, a Special Interest
  Group of the {ACL}}, pages 1631--1642. {ACL}.

\bibitem[{Sun et~al.(2019)Sun, Cheng, Gan, and Liu}]{Sun2019PKD}
Siqi Sun, Yu~Cheng, Zhe Gan, and Jingjing Liu. 2019.
\newblock \href {https://doi.org/10.18653/v1/D19-1441} {Patient knowledge
  distillation for {BERT} model compression}.
\newblock In \emph{Proceedings of the 2019 Conference on Empirical Methods in
  Natural Language Processing and the 9th International Joint Conference on
  Natural Language Processing, {EMNLP-IJCNLP} 2019, Hong Kong, China, November
  3-7, 2019}, pages 4322--4331. Association for Computational Linguistics.

\bibitem[{Sun et~al.(2021{\natexlab{a}})Sun, Liu, Qiu, and
  Huang}]{Sun2021Paradigm}
Tianxiang Sun, Xiangyang Liu, Xipeng Qiu, and Xuanjing Huang.
  2021{\natexlab{a}}.
\newblock \href {http://arxiv.org/abs/2109.12575} {Paradigm shift in natural
  language processing}.
\newblock \emph{CoRR}, abs/2109.12575.

\bibitem[{Sun et~al.(2020)Sun, Shao, Qiu, Guo, Hu, Huang, and
  Zhang}]{Sun2020Colake}
Tianxiang Sun, Yunfan Shao, Xipeng Qiu, Qipeng Guo, Yaru Hu, Xuanjing Huang,
  and Zheng Zhang. 2020.
\newblock \href {https://doi.org/10.18653/v1/2020.coling-main.327} {Colake:
  Contextualized language and knowledge embedding}.
\newblock In \emph{Proceedings of the 28th International Conference on
  Computational Linguistics, {COLING} 2020, Barcelona, Spain (Online), December
  8-13, 2020}, pages 3660--3670. International Committee on Computational
  Linguistics.

\bibitem[{Sun et~al.(2021{\natexlab{b}})Sun, Zhou, Liu, Zhang, Jiang, Cao,
  Huang, and Qiu}]{Sun2021Early}
Tianxiang Sun, Yunhua Zhou, Xiangyang Liu, Xinyu Zhang, Hao Jiang, Zhao Cao,
  Xuanjing Huang, and Xipeng Qiu. 2021{\natexlab{b}}.
\newblock \href {http://arxiv.org/abs/2105.13792} {Early exiting with ensemble
  internal classifiers}.
\newblock \emph{CoRR}, abs/2105.13792.

\bibitem[{Tay et~al.(2021)Tay, Dehghani, Abnar, Shen, Bahri, Pham, Rao, Yang,
  Ruder, and Metzler}]{Tay2021LRA}
Yi~Tay, Mostafa Dehghani, Samira Abnar, Yikang Shen, Dara Bahri, Philip Pham,
  Jinfeng Rao, Liu Yang, Sebastian Ruder, and Donald Metzler. 2021.
\newblock \href {https://openreview.net/forum?id=qVyeW-grC2k} {Long range arena
  : {A} benchmark for efficient transformers}.
\newblock In \emph{9th International Conference on Learning Representations,
  {ICLR} 2021, Virtual Event, Austria, May 3-7, 2021}. OpenReview.net.

\bibitem[{Turc et~al.(2019)Turc, Chang, Lee, and
  Toutanova}]{Turc2019BERTComplete}
Iulia Turc, Ming{-}Wei Chang, Kenton Lee, and Kristina Toutanova. 2019.
\newblock \href {http://arxiv.org/abs/1908.08962} {Well-read students learn
  better: The impact of student initialization on knowledge distillation}.
\newblock \emph{CoRR}, abs/1908.08962.

\bibitem[{Vaswani et~al.(2017)Vaswani, Shazeer, Parmar, Uszkoreit, Jones,
  Gomez, Kaiser, and Polosukhin}]{Vaswani2017Attention}
Ashish Vaswani, Noam Shazeer, Niki Parmar, Jakob Uszkoreit, Llion Jones,
  Aidan~N. Gomez, Lukasz Kaiser, and Illia Polosukhin. 2017.
\newblock \href
  {https://proceedings.neurips.cc/paper/2017/hash/3f5ee243547dee91fbd053c1c4a845aa-Abstract.html}
  {Attention is all you need}.
\newblock In \emph{Advances in Neural Information Processing Systems 30: Annual
  Conference on Neural Information Processing Systems 2017, December 4-9, 2017,
  Long Beach, CA, {USA}}, pages 5998--6008.

\bibitem[{Wang et~al.(2019{\natexlab{a}})Wang, Pruksachatkun, Nangia, Singh,
  Michael, Hill, Levy, and Bowman}]{Wang2019Superglue}
Alex Wang, Yada Pruksachatkun, Nikita Nangia, Amanpreet Singh, Julian Michael,
  Felix Hill, Omer Levy, and Samuel~R. Bowman. 2019{\natexlab{a}}.
\newblock \href
  {https://proceedings.neurips.cc/paper/2019/hash/4496bf24afe7fab6f046bf4923da8de6-Abstract.html}
  {Superglue: {A} stickier benchmark for general-purpose language understanding
  systems}.
\newblock In \emph{Advances in Neural Information Processing Systems 32: Annual
  Conference on Neural Information Processing Systems 2019, NeurIPS 2019,
  December 8-14, 2019, Vancouver, BC, Canada}, pages 3261--3275.

\bibitem[{Wang et~al.(2019{\natexlab{b}})Wang, Singh, Michael, Hill, Levy, and
  Bowman}]{Wang2019GLUE}
Alex Wang, Amanpreet Singh, Julian Michael, Felix Hill, Omer Levy, and
  Samuel~R. Bowman. 2019{\natexlab{b}}.
\newblock \href {https://openreview.net/forum?id=rJ4km2R5t7} {{GLUE:} {A}
  multi-task benchmark and analysis platform for natural language
  understanding}.
\newblock In \emph{7th International Conference on Learning Representations,
  {ICLR} 2019, New Orleans, LA, USA, May 6-9, 2019}. OpenReview.net.

\bibitem[{Williams et~al.(2018)Williams, Nangia, and Bowman}]{Williams2018MNLI}
Adina Williams, Nikita Nangia, and Samuel~R. Bowman. 2018.
\newblock \href {https://doi.org/10.18653/v1/n18-1101} {A broad-coverage
  challenge corpus for sentence understanding through inference}.
\newblock In \emph{Proceedings of the 2018 Conference of the North American
  Chapter of the Association for Computational Linguistics: Human Language
  Technologies, {NAACL-HLT} 2018, New Orleans, Louisiana, USA, June 1-6, 2018,
  Volume 1 (Long Papers)}, pages 1112--1122. Association for Computational
  Linguistics.

\bibitem[{Wolf et~al.(2020)Wolf, Debut, Sanh, Chaumond, Delangue, Moi, Cistac,
  Rault, Louf, Funtowicz, Davison, Shleifer, von Platen, Ma, Jernite, Plu, Xu,
  Scao, Gugger, Drame, Lhoest, and Rush}]{Wolf2020Transformers}
Thomas Wolf, Lysandre Debut, Victor Sanh, Julien Chaumond, Clement Delangue,
  Anthony Moi, Pierric Cistac, Tim Rault, R{\'{e}}mi Louf, Morgan Funtowicz,
  Joe Davison, Sam Shleifer, Patrick von Platen, Clara Ma, Yacine Jernite,
  Julien Plu, Canwen Xu, Teven~Le Scao, Sylvain Gugger, Mariama Drame, Quentin
  Lhoest, and Alexander~M. Rush. 2020.
\newblock \href {https://doi.org/10.18653/v1/2020.emnlp-demos.6} {Transformers:
  State-of-the-art natural language processing}.
\newblock In \emph{Proceedings of {EMNLP} 2020 - Demos}, pages 38--45.
  Association for Computational Linguistics.

\bibitem[{Xin et~al.(2020)Xin, Tang, Lee, Yu, and Lin}]{Xin2020DeeBERT}
Ji~Xin, Raphael Tang, Jaejun Lee, Yaoliang Yu, and Jimmy Lin. 2020.
\newblock \href {https://www.aclweb.org/anthology/2020.acl-main.204/} {Deebert:
  Dynamic early exiting for accelerating {BERT} inference}.
\newblock In \emph{Proceedings of the 58th Annual Meeting of the Association
  for Computational Linguistics, {ACL} 2020, Online, July 5-10, 2020}, pages
  2246--2251. Association for Computational Linguistics.

\bibitem[{Xin et~al.(2021)Xin, Tang, Yu, and Lin}]{Xin2021BERxiT}
Ji~Xin, Raphael Tang, Yaoliang Yu, and Jimmy Lin. 2021.
\newblock \href {https://aclanthology.org/2021.eacl-main.8/} {Berxit: Early
  exiting for {BERT} with better fine-tuning and extension to regression}.
\newblock In \emph{Proceedings of {EACL} 2021}, pages 91--104. Association for
  Computational Linguistics.

\bibitem[{Xu et~al.(2020{\natexlab{a}})Xu, Zhou, Ge, Wei, and
  Zhou}]{Xu2020BERTTheseus}
Canwen Xu, Wangchunshu Zhou, Tao Ge, Furu Wei, and Ming Zhou.
  2020{\natexlab{a}}.
\newblock \href {https://doi.org/10.18653/v1/2020.emnlp-main.633}
  {Bert-of-theseus: Compressing {BERT} by progressive module replacing}.
\newblock In \emph{Proceedings of the 2020 Conference on Empirical Methods in
  Natural Language Processing, {EMNLP} 2020, Online, November 16-20, 2020},
  pages 7859--7869. Association for Computational Linguistics.

\bibitem[{Xu et~al.(2020{\natexlab{b}})Xu, Hu, Zhang, Li, Cao, Li, Xu, Sun, Yu,
  Yu, Tian, Dong, Liu, Shi, Cui, Li, Zeng, Wang, Xie, Li, Patterson, Tian,
  Zhang, Zhou, Liu, Zhao, Zhao, Yue, Zhang, Yang, Richardson, and
  Lan}]{Xu2020CLUE}
Liang Xu, Hai Hu, Xuanwei Zhang, Lu~Li, Chenjie Cao, Yudong Li, Yechen Xu, Kai
  Sun, Dian Yu, Cong Yu, Yin Tian, Qianqian Dong, Weitang Liu, Bo~Shi, Yiming
  Cui, Junyi Li, Jun Zeng, Rongzhao Wang, Weijian Xie, Yanting Li, Yina
  Patterson, Zuoyu Tian, Yiwen Zhang, He~Zhou, Shaoweihua Liu, Zhe Zhao, Qipeng
  Zhao, Cong Yue, Xinrui Zhang, Zhengliang Yang, Kyle Richardson, and Zhenzhong
  Lan. 2020{\natexlab{b}}.
\newblock \href {https://doi.org/10.18653/v1/2020.coling-main.419} {{CLUE:} {A}
  chinese language understanding evaluation benchmark}.
\newblock In \emph{Proceedings of the 28th International Conference on
  Computational Linguistics, {COLING} 2020, Barcelona, Spain (Online), December
  8-13, 2020}, pages 4762--4772. International Committee on Computational
  Linguistics.

\bibitem[{Yang et~al.(2019)Yang, Dai, Yang, Carbonell, Salakhutdinov, and
  Le}]{Yang2019XLNet}
Zhilin Yang, Zihang Dai, Yiming Yang, Jaime~G. Carbonell, Ruslan Salakhutdinov,
  and Quoc~V. Le. 2019.
\newblock \href
  {http://papers.nips.cc/paper/8812-xlnet-generalized-autoregressive-pretraining-for-language-understanding}
  {Xlnet: Generalized autoregressive pretraining for language understanding}.
\newblock In \emph{Advances in Neural Information Processing Systems 32: Annual
  Conference on Neural Information Processing Systems 2019, NeurIPS 2019,
  December 8-14, 2019, Vancouver, BC, Canada}, pages 5754--5764.

\bibitem[{Yang et~al.(2018)Yang, Qi, Zhang, Bengio, Cohen, Salakhutdinov, and
  Manning}]{Yang2018HotpotQA}
Zhilin Yang, Peng Qi, Saizheng Zhang, Yoshua Bengio, William~W. Cohen, Ruslan
  Salakhutdinov, and Christopher~D. Manning. 2018.
\newblock \href {https://doi.org/10.18653/v1/d18-1259} {Hotpotqa: {A} dataset
  for diverse, explainable multi-hop question answering}.
\newblock In \emph{Proceedings of the 2018 Conference on Empirical Methods in
  Natural Language Processing, Brussels, Belgium, October 31 - November 4,
  2018}, pages 2369--2380. Association for Computational Linguistics.

\bibitem[{Zhou et~al.(2020{\natexlab{a}})Zhou, Xu, Ge, McAuley, Xu, and
  Wei}]{Zhou2020PABEE}
Wangchunshu Zhou, Canwen Xu, Tao Ge, Julian~J. McAuley, Ke~Xu, and Furu Wei.
  2020{\natexlab{a}}.
\newblock \href
  {https://proceedings.neurips.cc/paper/2020/hash/d4dd111a4fd973394238aca5c05bebe3-Abstract.html}
  {{BERT} loses patience: Fast and robust inference with early exit}.
\newblock In \emph{Advances in Neural Information Processing Systems 33: Annual
  Conference on Neural Information Processing Systems 2020, NeurIPS 2020,
  December 6-12, 2020, virtual}.

\bibitem[{Zhou et~al.(2020{\natexlab{b}})Zhou, Xu, Ge, McAuley, Xu, and
  Wei}]{Zhou2020BERT}
Wangchunshu Zhou, Canwen Xu, Tao Ge, Julian~J. McAuley, Ke~Xu, and Furu Wei.
  2020{\natexlab{b}}.
\newblock \href
  {https://proceedings.neurips.cc/paper/2020/hash/d4dd111a4fd973394238aca5c05bebe3-Abstract.html}
  {{BERT} loses patience: Fast and robust inference with early exit}.
\newblock In \emph{Advances in Neural Information Processing Systems 33: Annual
  Conference on Neural Information Processing Systems 2020, NeurIPS 2020,
  December 6-12, 2020, virtual}.

\bibitem[{Zhu(2021)}]{Zhu2021Leebert}
Wei Zhu. 2021.
\newblock \href {https://doi.org/10.18653/v1/2021.acl-long.231} {Leebert:
  Learned early exit for {BERT} with cross-level optimization}.
\newblock In \emph{Proceedings of the 59th Annual Meeting of the Association
  for Computational Linguistics and the 11th International Joint Conference on
  Natural Language Processing, {ACL/IJCNLP} 2021, (Volume 1: Long Papers),
  Virtual Event, August 1-6, 2021}, pages 2968--2980. Association for
  Computational Linguistics.

\bibitem[{Zhu et~al.(2015)Zhu, Kiros, Zemel, Salakhutdinov, Urtasun, Torralba,
  and Fidler}]{zhu2015bookcorpus}
Yukun Zhu, Ryan Kiros, Richard~S. Zemel, Ruslan Salakhutdinov, Raquel Urtasun,
  Antonio Torralba, and Sanja Fidler. 2015.
\newblock \href {https://doi.org/10.1109/ICCV.2015.11} {Aligning books and
  movies: Towards story-like visual explanations by watching movies and reading
  books}.
\newblock In \emph{2015 {IEEE} International Conference on Computer Vision,
  {ICCV} 2015, Santiago, Chile, December 7-13, 2015}, pages 19--27. {IEEE}
  Computer Society.

\end{thebibliography}
\bibliographystyle{acl_natbib}

\appendix
\section*{Appendix}
\section{Details of Evaluation}
\label{sec:appendix-eval}

\subsection{Submission and Evaluation}
\label{sec:eval}
ELUE supports two kinds of submissions: submitting test files, or submitting from a paper. 
\paragraph{Submit test files}
Users are required to submit two kinds of files: (1) predicted test files, and (2) a model definition file in Python. The predicted test files can be multiple, each indicates the prediction under a certain efficiency. The submitted test files should be in the following format:
\begin{table}[h!]
    \centering
    \resizebox{\linewidth}{!}{
    \begin{tabular}{|lll|}
    \hline
    \textcolor{teal}{index} & \textcolor{teal}{pred} & \textcolor{teal}{modules}                                                                  \\
    0     & 1          & (10),emb; (10,768),layer\_1; (768),exit\_1                                 \\
    1     & 0          & (15),emb; (15,768),layer\_1; (768),exit\_1; (15,768),layer\_2; (768),exit\_2 \\
    2     & 1          & (12),emb; (12,768),layer\_1; (768),exit\_1                               \\
    ...   & ...        & ...                                                                      \\ \hline
    \end{tabular}
    }
\end{table}
\\
Different from traditional predicted test files as in GLUE, an additional column "modules" is required to indicate the activated modules to predict each sample. The numbers before each module represent the input shape of that module, e.g. the "(10)" before "emb" indicates that the input of "emb" is a sequence of length 10. Note that this format is also compatible with token-level early exiting methods~\cite{Li2020Accelerating}, where the sequence length is progressively reduced as the processing of layers.

Along with the test files, a Python file to define the model is also required. Figure~\ref{fig:example_submission} is an example Python file using PyTorch~\cite{Paszke2019Pytorch} and Transformers~\cite{Wolf2020Transformers}.

With the submitted Python file, ELUE is able to evaluate the average FLOPs on a dataset, and the number of parameters of the model.

In cases that the evaluation is not applicable, e.g. the programming language, or dependencies of the submitted Python file is not supported in ELUE, the user is allowed to evaluate FLOPs and number of parameters by themselves and upload their results along with the predictions to the ELUE website.

\paragraph{Submit from a paper}
Inspired by Paper with Code\footnote{\href{https://paperswithcode.com/}{https://paperswithcode.com/}}, we also expect that ELUE can serve as an open-source platform that can facilitate future research. Therefore, there is a track for the authors of published papers to share their experimental results on ELUE datasets.

\paragraph{Performance Metrics}
Since the classes in MRPC are imbalanced, we report the unweighted average of accuracy and F1 score. For STS-B, we evaluate and report the Pearson and Spearman correlation coefficients. For other datasets, we simply adopt accuracy as the metric.

\begin{figure}[t]
    \centering
    {\tiny
\begin{minted}[frame=single]{Python}

# import packages
import torch.nn as nn
from transformers import BertConfig
...

# module definitions
class ElasticBERTEmbeddings(nn.Module):
    def __init__():
        ...
    def forward(x):
        ...

class ElasticBERTLayer(nn.Module):
    def __init__():
        ...
    def forward(x)
        ...

class ElasticBERT(nn.Module):
    def __init__():
        ...
    def forward(x)
        ...

# module dict
config = BertConfig(num_labels=2)
module_list = {
    'emb': ElasticBertEmbeddings(config),
    'layer_1': ElasticBertLayer(config),
    'exit_1': nn.Linear(config.hidden_size, num_labels),
    'layer_2': ElasticBertLayer(config),
    'exit_2': nn.Linear(config.hidden_size, num_labels),
    ...
}
entire_model = ElasticBERT(config)
\end{minted}
}
    \caption{An example Python file for submission.}
    \label{fig:example_submission}
\end{figure}

\section{Experimental Details and Additional Results}
\label{sec:appendix-elasticBERT}

\subsection{Details of Training ElasticBERT}\label{sec:trainingsetup}

The parameters of ElasticBERT are initialized with BERT, and therefore it has the same vocabulary and tokenizer as BERT. ElasticBERT is pre-trained on $\sim$160GB uncompressed English text corpora, which is comprised of English Wikipedia (12GB), BookCorpus (4GB)~\cite{zhu2015bookcorpus}, OpenWebText (38GB)~\cite{Gokaslan2019OpenWeb}, and part of the C4 corpus (110GB)~\cite{Raffel2020T5}. We use Adam optimizer~\cite{kingma2015adam} to pre-train ElasticBERT\textsubscript{BASE} and ElasticBERT\textsubscript{LARGE} and other hyperparameters are listed in Table \ref{tab:hyperparam for pretraining}. Our implementation is based on Huggingface's Transformers~\cite{Wolf2020Transformers} and the Megatron-LM toolkit~\cite{shoeybi2019megatron}. ElasticBERT is trained on 64 32G NVIDIA Tesla V100 GPUs.



\begin{table}[t!]
\centering
\resizebox{\linewidth}{!}{
\begin{tabular}{lcc}
\toprule
\textbf{Hyperparameter} & \textbf{ElasticBERT\textsubscript{BASE}} & \textbf{ElasticBERT\textsubscript{LARGE}} \\
\midrule
Adam $\beta_1$ & 0.9 & 0.9 \\
Adam $\beta_2$ & 0.999 & 0.999 \\
Peak Learning Rate & 2e-4 & 2e-4 \\
Warm Up Type & Linear & Linear \\
Warm Up Rate & 0.04 & 0.04 \\
Weight Decay & 0.01 & 0.01 \\
Batch Size & 4096 & 4096 \\
Training Steps & 125k & 125k \\
Number of Layers & 12 & 24 \\
Hidden Size & 768 & 1024 \\
Attention Heads & 12 & 16 \\
FFN Intermediate Size & 3072 & 4096 \\
\bottomrule
\end{tabular}
}
\caption{Hyperparameters for ElasticBERT pre-training}
\label{tab:hyperparam for pretraining}
\end{table}

\subsection{Evaluating ElasticBERT on GLUE}
\label{sec:elasticBERTonGLUE}
To verify the effectiveness and the elasticity of ElasticBERT, we also evaluate ElasticBERT and our static baselines on the GLUE benchmark. We evaluate the first 6/12 layers of the BASE models, and the first 6/24 layers of the LARGE models. 


Experimental results of ElasticBERT and our baseline models on GLUE are presented in Table~\ref{tab:glue_static}, from which we find that ElasticBERT outperforms BERT and ALBERT with the same number of layers, but is weaker than RoBERTa in the 12/24 layers configuration. Compared with ElasticBERT that is trained for 125K steps with batch size of 4K, RoBERTa is trained for 500K steps with batch size of 8K, which makes its number of training samples 8 times larger than that of ElasticBERT. When using fewer layers (6 layers of BASE and LARGE models), ElasticBERT achieves the best performance among the static baselines, confirming its great elasticity.

\begin{table}[h]
\centering
\small
\begin{tabular}{lc}
\toprule
\textbf{Grouping}                                                                        & \textbf{Accuracy} \\ \midrule
w/o Grouping                                                                             & \textbf{76.7}              \\ \midrule
\begin{tabular}[c]{@{}l@{}}$\mathcal{G}_1$=\{1, 3, 5, 7, 9, 11, 12\}\\ $\mathcal{G}_2$=\{2, 4, 6, 8, 10, 12\}\end{tabular}          & \textbf{76.7}              \\ \midrule
\begin{tabular}[c]{@{}l@{}}$\mathcal{G}_1$=\{1, 4, 7, 10, 12\}\\ $\mathcal{G}_2$=\{2, 5, 8, 11, 12\}\\ $\mathcal{G}_3$=\{3, 6, 9, 12\}\end{tabular} & 75.7              \\ \midrule
\begin{tabular}[c]{@{}l@{}}$\mathcal{G}_1$=\{1, 2, 3, 4, 12\}\\ $\mathcal{G}_2$=\{5, 6, 7, 8, 12\}\\ $\mathcal{G}_3$=\{9, 10, 11, 12\}\end{tabular} & 75.5              \\ \midrule
\begin{tabular}[c]{@{}l@{}}$\mathcal{G}_1$=\{1, 2, 3, 4, 5, 6, 12\}\\ $\mathcal{G}_2$=\{7, 8, 9, 10, 11, 12\}\end{tabular}          & 75.9              \\ \bottomrule
\end{tabular}
\caption{The average accuracy acorss all the BERT exits on the MNLI dataset with different grouping.}
\label{tab:grouping}
\end{table}

\begin{table*}[t!]
\centering
\resizebox{\linewidth}{!}{
\begin{tabular}{lccccccccccc}
\toprule
\textbf{Model} & \textbf{\#Params} & \textbf{\#FLOPs} & \textbf{CoLA} & \textbf{MNLI-m/mm} & \textbf{MRPC} & \textbf{QNLI} & \textbf{QQP} & \textbf{RTE} & \textbf{SST-2} & \textbf{STS-B} & \textbf{Average} \\
\midrule
\multicolumn{12}{c}{\textit{BASE Models}} \\
\midrule
BERT\textsubscript{BASE} & 109M & 6615M & 56.5 & 84.6/84.9 & 87.6 & 91.2 & 89.6 & 69.0 & 92.9 & 89.4 & 82.9 \\  
ALBERT\textsubscript{BASE} & 12M & 6861M & 56.8 & 84.9/85.6 & 90.5 & 91.4 & 89.2 & \textbf{78.3} & 92.8 & 90.7 & 84.5 \\
RoBERTa\textsubscript{BASE} & 125M & 6727M & 63.6 & \textbf{87.5/87.2} & 90.8 & \textbf{92.7} & \textbf{90.3} & 77.5 & \textbf{94.8} & \textbf{90.9} & \textbf{86.1} \\
LayerDrop\textsubscript{BASE} & 125M & 6727M & \textbf{64.5} & 86.4/86.5 & \textbf{91.6} & 92.2 & 89.9 & 71.1 & 93.7 & 88.6 & 84.9 \\

\textbf{ElasticBERT}\textsubscript{BASE} & 109M & 6615M & 64.3 & 85.3/85.9 & 91.0 & 92.0 & 90.2 & 76.5 & 94.3 & 90.7 & 85.6 \\
\midrule
BERT\textsubscript{BASE}-6L & 67M & 3308M & 44.6 & 81.4/81.4 & 84.9 & 87.4 & 88.7 & 65.7 & 90.9 & 88.1 & 79.2 \\  
ALBERT\textsubscript{BASE}-6L & 12M & 3435M & 52.4 & 82.6/82.2 & 89.0 & 89.8 & 88.7 & 70.4 & 90.8 & 89.6 & 81.7 \\
RoBERTa\textsubscript{BASE}-6L & 82M & 3364M & 44.4 & 84.2/\textbf{84.6} & 87.9 & 90.5 & \textbf{89.8} & 60.6 & 92.1 & 89.0 & 80.3 \\
LayerDrop\textsubscript{BASE}-6L & 82M & 3364M & 53.7 & 83.8/83.8 & 87.6 & 89.8 & 89.4 & 64.3 & 91.3 & 88.1 & 81.3 \\
HeadPrune-BERT\textsubscript{BASE} & 87M & 4744M & 48.7 & 71.0/79.7 & 80.2 & 86.1 & 84.7 & 62.5 & 89.4 & 85.2 & 76.4 \\
DistilBERT & 67M & 3308M & \textbf{55.6} & 82.1/82.0 & 86.5 & 89.2 & 88.8 & 63.9 & 91.3 & 86.7 & 80.7 \\
TinyBERT-6L & 67M & 3308M & 46.3 & 83.6/83.8 & 88.7 & 90.6 & 89.1 & 73.6 & 92.0 & 89.4 & 81.9 \\
BERT-of-Theseus & 67M & 3308M & 45.1 & 81.4/81.9 & 88.1 & 88.1 & 88.9 & 70.1 & 91.4 & 88.8 & 80.4 \\
\textbf{ElasticBERT}\textsubscript{BASE}-6L & 67M & 3308M & 53.7 & \textbf{84.3}/84.2 & \textbf{89.7} & \textbf{90.8} & 89.7 & \textbf{74.0} & \textbf{92.7} & \textbf{90.2} & \textbf{83.3} \\
\midrule
\multicolumn{12}{c}{\textit{Test Set Results}} \\
\midrule
TinyBERT-6L & 67M & 3308M & 42.5 & 83.2/82.4 & 86.2 & 89.6 & 79.6 & \textbf{73.0} & 91.8 & 85.7 & 79.3 \\
\textbf{ElasticBERT}\textsubscript{BASE}-6L & 67M & 3308M & \textbf{49.1} & \textbf{83.7/83.4} & \textbf{87.3} & \textbf{90.4} & \textbf{79.7} & 68.7 & \textbf{92.9} & \textbf{86.9} & \textbf{80.3} \\
\midrule
\multicolumn{12}{c}{\textit{LARGE Models}} \\
\midrule
BERT\textsubscript{LARGE} & 335M & 23446M & 61.6 & 86.2/86 & 90.1 & 92.2 & 90.1 & 72.9 & 93.5 & 90.4 & 84.8 \\  
ALBERT\textsubscript{LARGE} & 18M & 24296M & 60.1 & 86/86.1 & 90.4 & 91.6 & 89.6 & 83.0 & 95.2 & 91.4 & 85.9 \\
RoBERTa\textsubscript{LARGE} & 355M & 23840M & 66.4 & 89/89.6 & 91.6 & \textbf{94.2} & 90.7 & 86.6 & 95.4 & 92.3 & \textbf{88.4} \\
LayerDrop\textsubscript{LARGE} & 355M & 23840M & \textbf{66.6} & \textbf{89.7/89.6} & 91.2 & 93.9 & 88.5 & \textbf{86.6} & \textbf{95.5} & \textbf{92.6} & 88.2 \\
\textbf{ElasticBERT}\textsubscript{LARGE} & 335M & 23446M & 66.3 & 88/88.5 & \textbf{92.0} & 93.6 & \textbf{90.9} & 83.1 & 95.3 & 91.7 & 87.7 \\
\midrule
BERT\textsubscript{LARGE}-6L & 108M & 5863M & 20.2 & 76.5/76.5 & 76.4 & 84.3 & 87.3 & 58.5 & 89.7 & 77.3 & 78.5 \\  
ALBERT\textsubscript{LARGE}-6L & 18M & 6083M & 51.7 & 82.2/82.9 & 86.5 & 89.4 & 88.6 & 66.4 & \textbf{92.2} & 89.4 & 81.0 \\
RoBERTa\textsubscript{LARGE}-6L & 129M & 5962M & 43.3 & 80.4/80.9 & 80.0 & 86.1 & 88.9 & 54.9 & 90.1 & 80.5 & 76.1 \\
LayerDrop\textsubscript{LARGE}-6L & 129M & 5962M & 44.3 & 81.4/81.0 & 79.8 & 87.1 & 88.5 & 53.1 & 91.4 & 83.0 & 76.6 \\
\textbf{ElasticBERT}\textsubscript{LARGE}-6L & 108M & 5863M & \textbf{53.9} & \textbf{83.5/84.3} & \textbf{89.6} & \textbf{90.8} & \textbf{90.1} & \textbf{71.1} & 91.9 & \textbf{90.1} & \textbf{82.8} \\
\midrule
\multicolumn{12}{c}{\textit{Test Set Results}} \\
\midrule
ALBERT\textsubscript{LARGE}-6L & 18M & 6083M & 46.5 & 81.9/82.2 & 84.7 & 88.5 & 78.9 & 62.3 & 91.3 & 85.1 & 77.9 \\
\textbf{ElasticBERT}\textsubscript{LARGE}-6L & 108M & 5863M & \textbf{47.2} & \textbf{83.2/82.6} & \textbf{86.2} & \textbf{90.4} & \textbf{80.2} & \textbf{67.0} & \textbf{92.5} & \textbf{86.3} & \textbf{79.5} \\
\midrule
\bottomrule
\end{tabular}
}
\caption{ElasticBERT and static baseline performance on GLUE tasks. For MRPC, we report the mean of accuracy and F1. For STS-B, we report Pearson and Spearman correlation. For CoLA, we report Matthews correlation. For all other tasks we report accuracy.\vspace{-0.5cm}}
\label{tab:glue_static}
\end{table*}

\subsection{Ablation Study}
\label{sec:ablation}
\paragraph{About the Training Strategy}
ElasticBERT adopts the gradient equilibrium (GE) to alleviate the conflict between the losses at different exits. Here, we compare GE with two other existing training strategies, two-stage training~\cite{Xin2020DeeBERT} and weighted training~\cite{Zhou2020PABEE}. Two-stage training is that, first training the top classifier along with the backbone model, and then freeze the parameters of the backbone model and train the injected internal classifiers. By this, two-stage training maintains the performance of the top classifier. Weighted training is to weight the loss of each exit according to the corresponding layer, which is
\begin{equation}
    \mathcal{L} = \frac{\sum_{l=1}^L l\cdot \mathcal{L}_l}{\sum_{l=1}^{L} l}.
\end{equation}

\begin{figure}[t]
    \centering
    \begin{subfigure}{.48\linewidth}
    \centering
    \includegraphics[width=\linewidth]{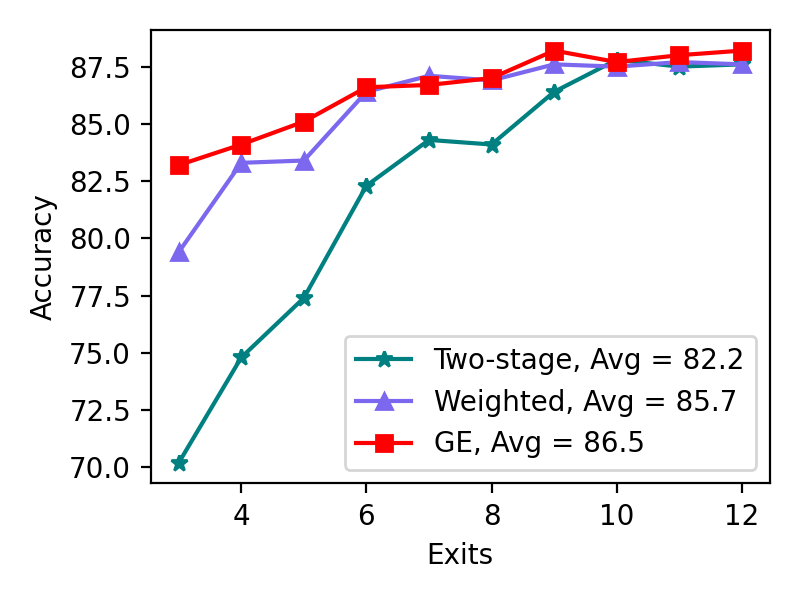}
    \caption{SST-2}
    \end{subfigure}
    \begin{subfigure}{.48\linewidth}
    \centering
    \includegraphics[width=\linewidth]{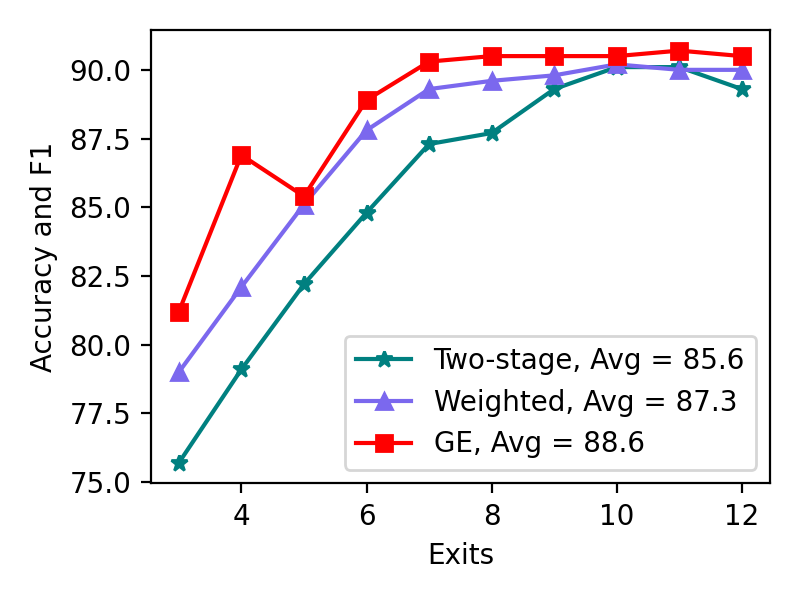}
    \caption{MRPC}
    \end{subfigure}
    \caption{Performance of the ElasticBERT exits at different layers with different training strategies.\vspace{-0.5cm}}
    \label{fig:ablation_training}
\end{figure}

Experimental results of ElasticBERT with the three training strategies are shown in Figure~\ref{fig:ablation_training}. It can be observed that training with GE strategy performs the best on both SST-2 and MRPC.

\begin{table*}[t]
    \centering
    \resizebox{0.8\linewidth}{!}{
    \begin{tabular}{lccccc}
    \toprule
        \textbf{Method} & \textbf{MNLI-m/mm} & \textbf{MRPC} & \textbf{QNLI} & \textbf{QQP} & \textbf{Average}  \\
        \midrule
        \multicolumn{6}{c}{\textit{12 Layers}} \\
        \midrule
         ElasticBERT\textsubscript{BASE} & 85.4/\textbf{85.7} & 89.2 & 91.9 & \textbf{89.8} & \textbf{88.4} \\
         \ \ w/o Grouped Training & 85.3/85.1 & \textbf{89.3} & \textbf{92.0} & 89.8 & 88.3 \\
         \ \ w/o Grouped Training + GE & \textbf{85.5}/85.6 & 89.0 & 91.8 & 89.5 & 88.3 \\
        \midrule
        \multicolumn{6}{c}{\textit{6 Layers}} \\
        \midrule
         ElasticBERT\textsubscript{BASE} & 83.9/\textbf{83.6} & 87.1 & 90.6 & \textbf{89.6} & 87.0 \\
         \ \ w/o Grouped Training & \textbf{84.5}/83.6 & \textbf{87.5} & \textbf{90.7} & 89.4 & \textbf{87.1} \\
         \ \ w/o Grouped Training + GE & 83.5/83.3 & 87.2 & 90.7 & 88.3 & 86.6 \\
    \bottomrule
    \end{tabular}
}
    \caption{Comparison between ElasticBERT\textsubscript{BASE} with and without GE and Grouped Training. Here, ElastiBERT\textsubscript{BASE} is pre-trained using Wikipeida and BookCorpus data. ElasticBERT\textsubscript{BASE} denotes that model is pre-trained with GE and Grouped Training.}
    \label{tab:comparison between with and without GE and Grouped Training}
\end{table*}

\begin{table}[h]
    \centering
    \resizebox{.9\linewidth}{!}{
    \begin{tabular}{lc}
    \toprule
        \textbf{Method} & \textbf{Training Time (h)}  \\
        \midrule
         ElasticBERT\textsubscript{BASE} & 106.0 \\
         -w/o Grouped Training & 186.0 \\
         -w/o Grouped Training + GE & 174.5 \\
    \bottomrule
    \end{tabular}
    }
    \caption{The training time for different training strategies. All models are trained on the same GPU servers.\vspace{-0.5cm}}
    \label{tab:pre-training time}
\end{table}

\paragraph{About the Grouped Exits}
As shown in Eq. (\ref{eq:group}), we divide the $L$ exits into different groups to speedup the pre-training. Therefore, how to group these exits needs to be explored. Here we evaluate four different grouping methods, as described in Table~\ref{tab:grouping}. To keep the overall performance of the entire model, the exit classifier on the top of the model is included in each group. According to the experimental results in Table~\ref{tab:grouping}, we choose the odd/even grouping method for ElasticBERT\textsubscript{BASE}. Similarly, our experiments demonstrate that grouping 24 exits into $\mathcal{G}_1$=\{1, 4, 7, ..., 22, 24\}, $\mathcal{G}_2$=\{2, 5, 8, ..., 23, 24\}, and $\mathcal{G}_3$=\{3, 6, 9, ..., 21, 24\} works well for ElasticBERT\textsubscript{LARGE}.

\paragraph{Effect of Gradient Equilibrium}
To verify that GE can enhance performance, we pre-train ElasticBERT\textsubscript{BASE} with and without GE using Wikipedia and BookCorpus data. As shown in Table~\ref{tab:comparison between with and without GE and Grouped Training}, ElasticBERT with GE outperforms that without GE in two different configurations.

\paragraph{Effect of Grouped Training}
If we divide $L$ exits into $G$ groups, the number of samples used for training the remaining exits is $1/G$ of the last exit. To verify that this does not degrade the performance of the internal exits, we compare the performances of pre-training ElasticBERT\textsubscript{BASE} with and without Grouped Training. As shown in Table~\ref{tab:comparison between with and without GE and Grouped Training}, we can observe that ElasticBERT with Grouped Training and GE does not suffer much performance loss compared with that training with only GE. In addition, as shown in the Table~\ref{tab:pre-training time}, using Grouped Training reduces $\sim$43\% training time compared with that without Grouped Training.

\begin{figure}[t!]
    \centering
    \begin{subfigure}{.48\linewidth}
    \centering
    \includegraphics[width=\linewidth]{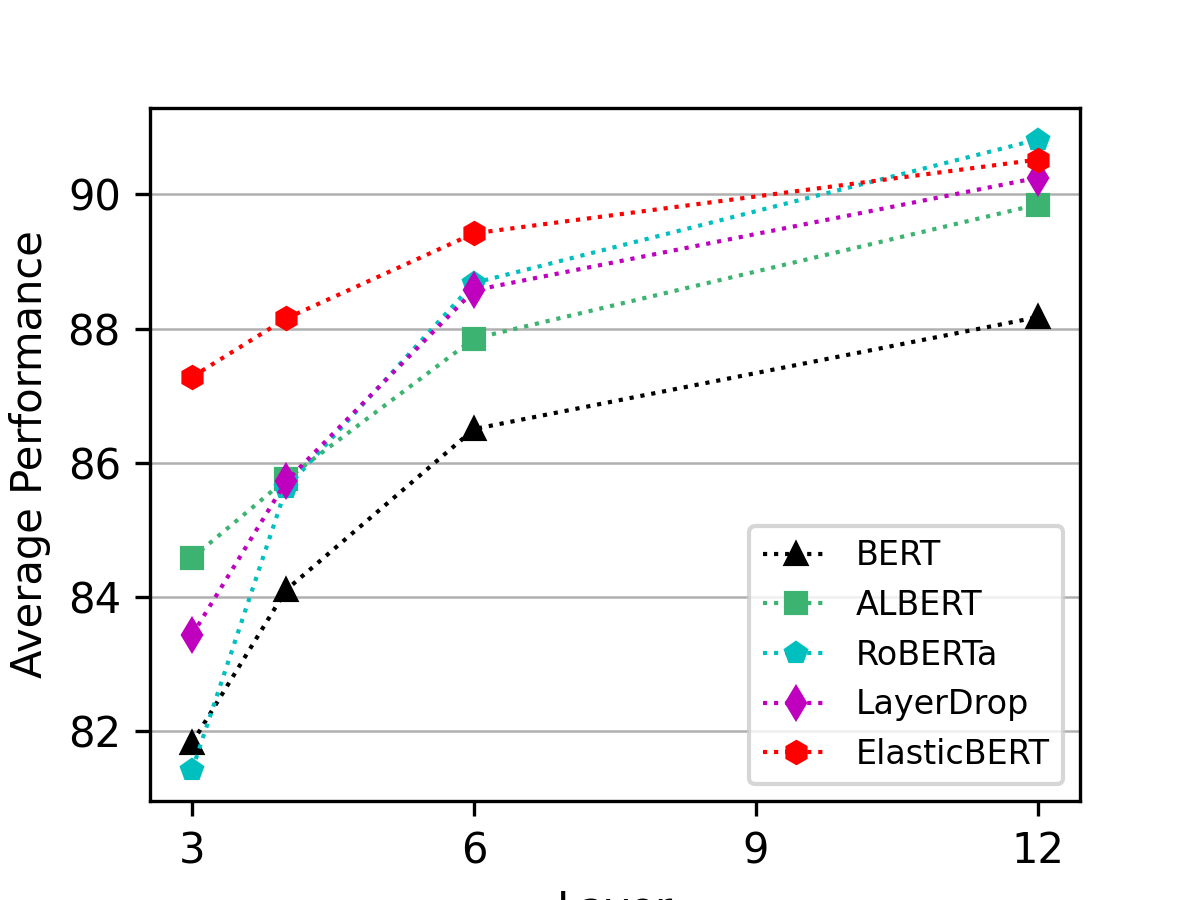}
    \caption{ElasticBERT\textsubscript{BASE}}
    \end{subfigure}
    \begin{subfigure}{.48\linewidth}
    \centering
    \includegraphics[width=\linewidth]{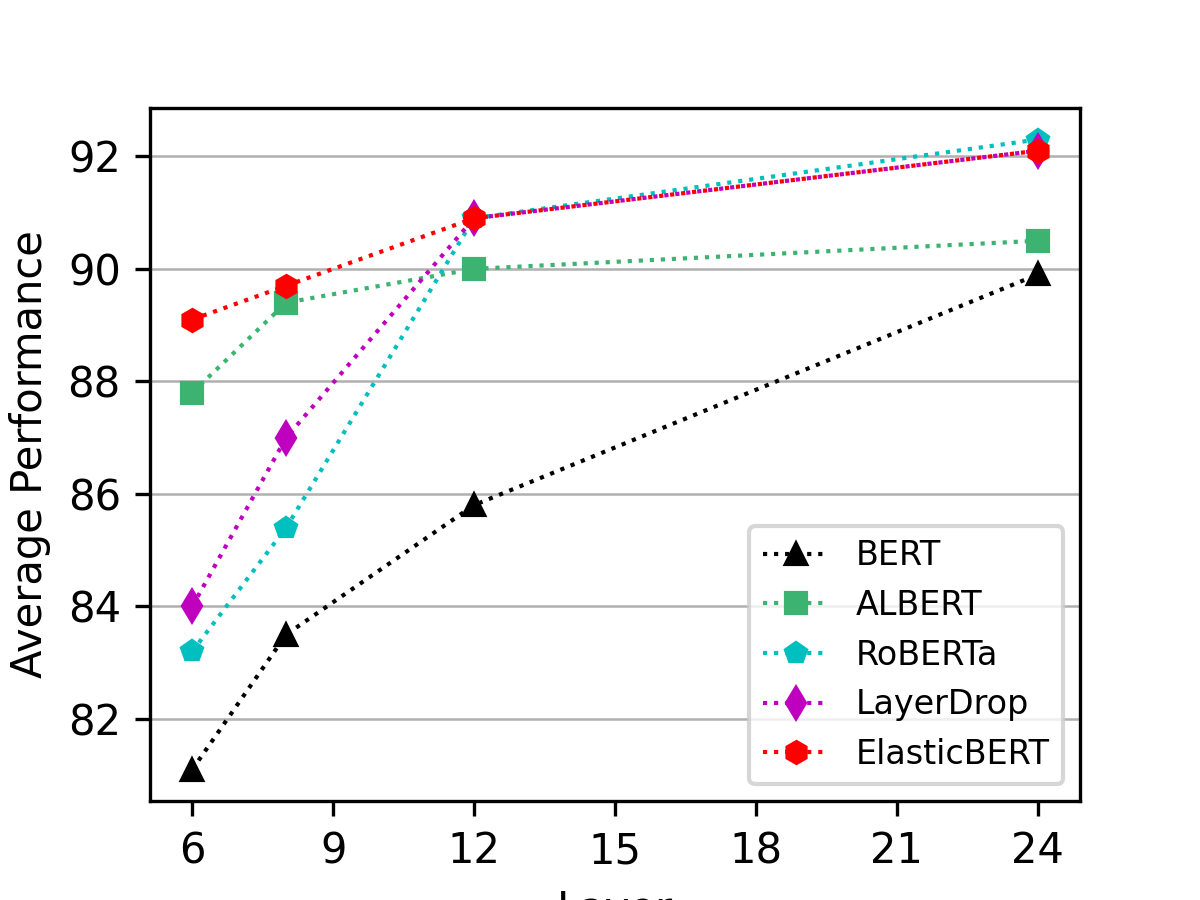}
    \caption{ElasticBERT\textsubscript{LARGE}}
    \end{subfigure}
    \caption{Comparison of average performance on ELUE test sets between the ElasticBERT and other pre-trained models.\vspace{-0.5cm}}
    \label{fig:overall comparison}
\end{figure}

\subsection{Overall Comparison}
We compare ElastiBERT with other large scale pre-trained models in Figure~\ref{fig:overall comparison}, from which we find that ElasticBERT is more robust to depth reduction. As the number of layers decreases, ElasticBERT offers greater advantages over other pre-trained models.

\section{ELUE Website}
The ELUE website is built using Vue and Spring Boot. We use MySQL for data storage and our private servers to run the scoring script for each submission. 




\end{document}